\documentclass[journal]{IEEEtran}
\usepackage{times}
\usepackage{amsmath,amsfonts}
\usepackage{algorithmic}
\usepackage{algorithm}
\usepackage{array}
\usepackage[caption=false,font=normalsize,labelfont=sf,textfont=sf]{subfig}
\usepackage{textcomp}
\usepackage{stfloats}
\usepackage{url}
\usepackage{verbatim}
\usepackage{graphicx}
\usepackage{cite}
\usepackage{makecell}
\usepackage{multirow}
\usepackage{tabularx}
\usepackage{booktabs} 
\usepackage{lipsum}
\usepackage{amssymb}

\begin{document}

\title{Occlusion-Aware Diffusion Model for Pedestrian Intention Prediction}

\author{ Yu Liu, Zhijie Liu, Zedong Yang, You-Fu Li,~\IEEEmembership{Fellow,~IEEE,} and He Kong,~\IEEEmembership{Member,~IEEE}
  
\thanks{This manuscript has been accepted to the IEEE Transactions on Intelligent Transportation Systems as a regular paper. This work was supported by the National Key R$\&$D Program of China under Grant No. 2024YFB4710902, the National Natural Science Foundation of China (NSFC) under Grant No. U24A20265, the Science, Technology, and Innovation Commission of Shenzhen Municipality, China, under Grant No. ZDSYS20220330161800001, JCYJ20240813094212017, the Shenzhen Science and Technology Program under Grant No. KQTD20221101093557010, the Guangdong Science and Technology Program under Grant No. 2024B1212010002. (Corresponding author: He Kong)}

\thanks{Yu Liu, Zhijie Liu, Zedong Yang, and He Kong are with the Shenzhen Key Laboratory of Control Theory and Intelligent Systems, Southern University of Science and Technology (SUSTech), Shenzhen 518055, China. Yu Liu is also with the Department of Mechanical Engineering, City University of Hong Kong, Hong Kong SAR, China. Emails: yuliu254-c@my.cityu.edu.hk; 12332642@mail.sustech.edu.cn; 12210810@mail.sustech.edu.cn; kongh@sustech.edu.cn. You-Fu Li is with the Department of Mechanical Engineering, City University of Hong Kong, Hong Kong SAR, China. Email: meyfli@cityu.edu.hk.}
}



\maketitle

\begin{abstract}
Predicting pedestrian crossing intentions is crucial for the navigation of mobile robots and intelligent vehicles. Although recent deep learning-based models have shown significant success in forecasting intentions, few consider incomplete observation under occlusion scenarios. To tackle this challenge, we propose an Occlusion-Aware Diffusion Model (ODM) that reconstructs occluded motion patterns and leverages them to guide future intention prediction. During the denoising stage, we introduce an occlusion-aware diffusion transformer architecture to estimate noise features associated with occluded patterns, thereby enhancing the model's ability to capture contextual relationships in occluded semantic scenarios. Furthermore, an occlusion mask-guided reverse process is introduced to effectively utilize observation information, reducing the accumulation of prediction errors and enhancing the accuracy of reconstructed motion features. The performance of the proposed method under various occlusion scenarios is comprehensively evaluated and compared with existing methods on popular benchmarks, namely PIE and JAAD. Extensive experimental results demonstrate that the proposed method achieves more robust performance than existing methods in the literature.

\end{abstract}


\section{Introduction}
\IEEEPARstart{W}{ith} the rapid advancement of intelligent sensing and computing technologies, much progress has been made in recent years in developing autonomous vehicles to enhance traffic efficiency and road safety. To prevent collisions, path planning of autonomous vehicles \cite{ref3,ref94} is essential, requiring an understanding of interactions between road users and the ability to forecast their potential actions \cite{ref2, ref1, ref4}.  

\begin{figure}[t]
\centering
\includegraphics[scale= 0.60]{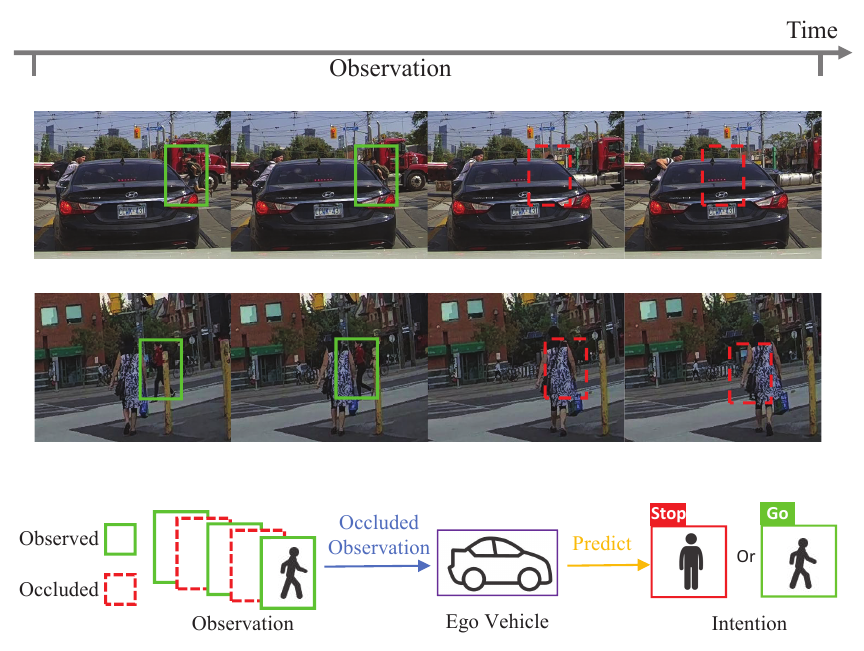}
\caption{The typical scenario of visual occlusion is illustrated here. Solid green lines represent the parts of the observation that are within the field of view and visible, while dashed red lines indicate positional features that are undetectable due to occlusion. Observations with occlusion pose a significant challenge for pedestrian intention prediction. }
\centering
\label{intro1}
\vspace{-1em}
\end{figure}

Given the observed traffic environment information and the historical motion data of individuals, pedestrian intention refers specifically to the binary decision of whether a pedestrian intends to cross the road within a short future time. One of the key challenges in intention prediction is the probabilistic and multimodal nature of road users' behaviors, which are influenced by a variety of factors \cite{ref27}. Recently, several methods have been proposed to incorporate multimodal data to estimate pedestrian intention, including, for example, visual context information \cite{ref9, ref10}, ego vehicle velocity \cite{ref11}, bounding box annotation \cite{ref73, ref14}, and pose information \cite{ref12}.

\IEEEpubidadjcol  

However, visual occlusions, which frequently occur in practical applications can cause onboard sensors to fail in detecting pedestrians' current states in complex environments, as illustrated in Figure \ref{intro1}. The inaccessibility of certain modalities may prevent vehicles from accurately estimating pedestrians' future motion intentions, posing a risk to their safe operation. Existing intention prediction methods typically rely on complete multimodal observation sequences and do not consider occluded observations. While a few motion prediction studies \cite{ref7, ref8} consider incomplete observations, they primarily focus on observable sequences and overlook the impact of missing motion patterns. To the best of our knowledge, learning-based prediction of pedestrian crossing intentions in occlusion scenarios has not been well-explored.

From this perspective, in this work, we propose an Occlusion-Aware Diffusion Model for Intention Prediction to address the challenges of intention prediction in occluded scenarios. Our framework integrates an occlusion-masked diffusion transformer with an occlusion mask-guided inference process. By leveraging multimodal inputs, such as bounding boxes and ego-vehicle velocity, the model reconstructs occluded motion features before estimating pedestrian crossing intentions. The occlusion masking strategy is designed to handle incomplete pedestrian motion observations caused by occlusion in traffic environments. Unlike token masking in NLP or conventional diffusion models, our approach directly targets motion features, enabling the diffusion model to recover and learn from occluded patterns for more accurate intention prediction. This method effectively captures both motion and occlusion patterns, allowing the model to reconstruct missing features while predicting future intentions. The contributions of this paper are as follows:

(1) We present a diffusion-based framework tailored for pedestrian intention prediction under occlusion scenarios. Unlike most existing studies that implicitly assume full visibility of pedestrian motion cues, our framework explicitly addresses the challenge of incomplete observations caused by occlusions, a problem insufficiently emphasized in prior work.

(2) We propose two key technical modules: the occlusion-masked diffusion transformer and the occlusion mask-guided reverse process. The occlusion-masked diffusion transformer that embeds semantic occlusion information into the denoising process, along with an occlusion mask-guided reverse process that leverages partial observations to mitigate error. Together, these designs enable more accurate reconstruction of motion features and intention prediction under occlusion. 

(3) Extensive experiments have been conducted to thoroughly evaluate the performance of the proposed approach. The experimental results on the PIE and JAAD datasets demonstrate that our method consistently achieves improved performance in terms of intention prediction under occlusion scenarios compared to state-of-the-art approaches \cite{ref90, ref74}.

The remainder of this article is organized as follows: Section \uppercase\expandafter{\romannumeral 2} reviews related methods for intention prediction and diffusion models. Section \uppercase\expandafter{\romannumeral 3} outlines the proposed methodology. Section \uppercase\expandafter{\romannumeral 4} describes the experimental setup and presents the evaluation results. Section \uppercase\expandafter{\romannumeral 5} concludes the paper.

\section{Related Works}

\subsection{Pedestrian Intention Prediction}
For intention prediction, early methods relied on dynamic physical cues, such as walking trajectories \cite{ref60}, trajectory endpoints \cite{ref61}, or pedestrian head orientations \cite{ref62}, to estimate crossing intentions. However, the limited and often simplistic incorporation of contextual features made these models vulnerable to challenges in complex, dynamic environments.

More recently, with the advent of benchmark datasets \cite{ref63, ref64} for pedestrian intention prediction, data-driven models have become a dominant approach in this field. For instance, multiple Recurrent Neural Network (RNN) branches have been employed to process different feature sources \cite{ref92}. In SFRNN \cite{ref91}, a Stacked RNN is used to hierarchically process features, gradually fusing them at each level to capture increasingly complex representations. PCPA \cite{ref90} proposed a Convolutional Neural Network (CNN) to extract visual features from scenes and RNNs to process non-image features for predicting crossing intentions. Inspired by the role of human pose in motion prediction tasks, \cite{ref65} integrated image-based 2D human pose features into the model to estimate crossing intentions. Similarly, pedestrian pose features were employed in \cite{ref66} to predict pedestrian crossing behaviors at intersections.

To better capture spatial interactions, Graph Neural Networks (GNNs) \cite{ref68} have been introduced to address intention prediction tasks by leveraging the spatiotemporal context of scenes. GNNs model interactions within scenarios using a topological graph structure, where dynamic objects, such as pedestrians and vehicles, are represented as graph nodes, and their interactions are captured as graph edges through adjacency matrices. For instance, ST CrossingPose \cite{ref69} proposed predicting pedestrian crossing intentions using spatial-temporal graph convolutional networks applied to skeleton data, effectively learning both spatial and temporal patterns. Additionally, Pedestrian Graph \cite{ref70} incorporated two convolutional modules to provide supplementary contextual information to the primary graph convolutional module, thereby enhancing prediction accuracy. 

Furthermore, the attention mechanism has been effectively utilized to capture semantic relationships within the data. For example, \cite{ref71} integrates attention with a multimodal prediction algorithm that combines various environmental information sources to forecast pedestrians' future crossing actions. Similarly, \cite{ref72} proposes an architecture that fuses different spatiotemporal features, thereby significantly improving the prediction of pedestrian crossing intentions.

Recently, the transformer model \cite{ref6}, originally developed for natural language processing (NLP), has significantly advanced the modelling of sequential data. Therefore, several Transformer-based models have been proposed for pedestrian intention prediction. For instance, CP2A \cite{ref73} presents a framework that employs multiple variations of the transformer model, using only bounding boxes as input features while achieving competitive performance. Similarly, TrEP \cite{ref74} introduces a transformer module integrated with a deep evidential learning model to address uncertainty in complex scenes. Meanwhile, PedCMT \cite{ref75} proposes a cross-modal Transformer-based intention prediction model that incorporates an uncertainty-aware multi-task learning approach to jointly predict crossing intentions and future bounding boxes.

However, in scenarios involving observation occlusion, where a pedestrian's position is intermittently unavailable, prediction models struggle with the loss of crucial motion patterns. This gap introduces significant uncertainty during both the feature extraction and prediction stages. Despite recent advancements, most existing intention prediction frameworks are not designed to consider such problems.

\subsection{Diffusion Probabilistic Model}
Inspired by non-equilibrium thermodynamics, \cite{ref80} first introduced the diffusion model, where a parameterized Markov chain transforms an initial noisy distribution into the target data distribution. As a branch of score-based generative frameworks, diffusion models \cite{ref81, ref82, ref95} have emerged as a prominent research focus. To alleviate the high computational cost of training and inference, \cite{ref83} enhances the approximation of maximum likelihood training by incorporating a learned covariance method and optimized noise schedule. \cite{ref88} proposes classifier-free guidance to refine the sampling strategy, while LDMs \cite{ref89} conducts generative learning in the latent space instead of the raw data space to reduce computational costs. Moreover, the widespread applicability of transformer networks has led to attempts to incorporate transformers into diffusion models. U-ViT \cite{ref84} replaces U-Net with a transformer model, employing long skip connections between shallow and deep layers. DiTs \cite{ref85} trains latent diffusion models with transformers and demonstrates their scalability with large model sizes and high feature resolutions. An asymmetric masking diffusion transformer is proposed in MDT \cite{ref86} to predict masked tokens from unmasked ones, facilitating the learning of relationships within an image.

In the context of motion prediction, diffusion models have recently emerged as one of the most prevalent tools for generating future motion. MID \cite{ref76} proposes a framework that formulates pedestrian trajectory prediction as a reverse process of motion indeterminacy diffusion, where contextual information is encoded to reduce uncertainty. EquiDiff \cite{ref77} employs recurrent neural networks and graph attention to model social interactions and estimate noise, with an SO(2)-equivariant transformer as the backbone model. By combining multimodal semantic map information with vehicle interaction dynamics, \cite{ref78} employs a masking mechanism to fully exploit historical data, while MotionDiffuser \cite{ref79} utilizes a controlled and guided sampling approach to forecast future motion. DICE \cite{ref87} introduces an efficient sampling mechanism that allows the model to maximize the number of sampled trajectories for improved accuracy while ensuring real-time inference. Additionally, it proposes a scoring mechanism to rank and select the most plausible trajectories.

In contrast to prior diffusion-based trajectory generation approaches, our method explicitly incorporates occlusion handling. We introduce an occlusion masking mechanism in the reverse process to mitigate noise and improve convergence, and design an occlusion-masked diffusion transformer that embeds occlusion semantics into motion reasoning. These additions enable the model to capture pedestrian motion dynamics under occlusion, providing a targeted enhancement over general diffusion prediction methods.

\section{METHODOLOGY}
\subsection{Preliminaries}
 
\begin{figure*}
\centering
    \includegraphics[scale= 0.62]{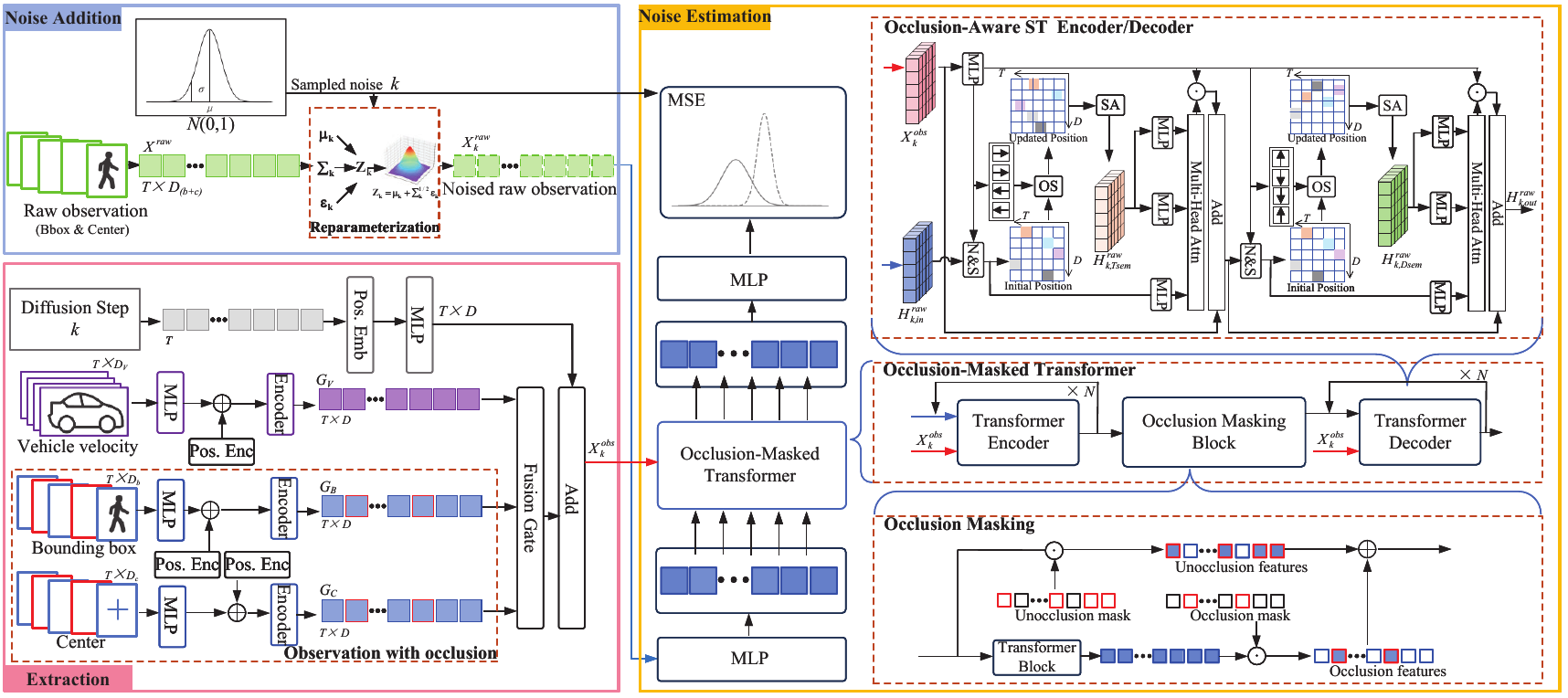}
    \caption{The training architecture of the noise estimation process consists of three blocks. In the noise addition block, raw observation sequences are corrupted with the noise of specific density, scaled by the diffusion step \(k\), to generate the noised feature \(X_{k}^{raw}\). In the observation extraction block, occluded pedestrian observations are integrated with the ego-vehicle's speed through a gating mechanism. These combined features are then added to the features from step \(k\) to form the observation vector \(X_{k}^{obs}\). In the noise estimation block, an occlusion-masked transformer is employed to predict the noise added at step \(k\), enhancing the model’s ability to learn semantic relationships. $\text{SA}$ is sampling operation, $\text{OS}$ indicates offset operation, $\text{N\&S}$ is normalization with scale and shift operation, $\odot$ is element-wise multiplication, and $\oplus$ is element-wise addition.
}
\centering
\label{figure structure2}
\vspace{-1em}
\end{figure*}

We briefly introduce the diffusion process \cite{ref81, ref83}, which is employed to reconstruct occluded motion features. The model consists of a forward and reverse Markov chain. In the forward process, noise is gradually added to raw sequences $X^{raw} \in \mathbb{R}^{T \times D}$ over $K$ steps following a scaled schedule, as shown in the Noise Addition block of Figure \ref{figure structure2}. The posterior is modeled as a first-order Markov chain with Gaussian transitions, and the distribution from $X^{raw}_{0}$ to $X^{raw}_{K}$ is given:
\begin{equation}
    \begin{array}{cc}
        q(X^{raw}_{1:K}{\mid}X^{raw}_{0} ) = \prod_{k=1}^{K} q(X^{raw}_{k}{\mid}X^{raw}_{k-1} ).
    \end{array}
\end{equation}

The posterior distribution \(X^{raw}_{k}\), denoting \(X^{raw}\) at step \(k\) conditioned on \(X^{raw}_{k-1}\), is formulated as follows:
\begin{equation}
    q(X^{raw}_{k}{\mid}X^{raw}_{k-1} ) = \mathcal{N}(X^{raw}_{k}; \sqrt{1-\beta_{k}}X^{raw}_{k-1}, \beta_{k}I), 
\end{equation}
where \(\beta_{k} \in (0,1)\) is the rescaled variance schedule controlling noise addition. The diffusion state at step $k$ is derived from the original state \(X_{0}^{raw}\) using the reparameterization method:
\begin{equation}
    \begin{array}{cc}
        q(X^{raw}_{k}{\mid}X^{raw}_{0} ) = \mathcal{N}(X^{raw}_{k}; \sqrt{\bar{\alpha}_{k}}X^{raw}_{0}, (1-{\bar{\alpha}_{k}})I), 
    \end{array}
    \label{eq3}
\end{equation}
\begin{equation}
    \begin{array}{cc}
        X^{raw}_{k} = \sqrt{\bar{\alpha}_{k}}X^{raw}_{0} + \sqrt{(1-{\bar{\alpha}_{k}})}\epsilon,\  \epsilon \sim \mathcal{N}(0, I),
    \end{array}
\end{equation}
where \(\alpha_{k} = 1 - \beta_{k}\), \(\bar{\alpha}_{k} = \prod_{i=1}^{k}\alpha_{i}\), and \(\epsilon\) is noise sampled from a standard Gaussian. As $k$ grows, $X^{raw}_{k}$ approximates the Gaussian latent space $\mathcal{N}(0, I)$.

In the reverse process, occluded features $X^{obs}_{k} \in \mathbb{R}^{T \times D}$ serve as input, and a parameterized Gaussian transition denoises them. Unlike the forward process, it removes noise step by step to recover the raw distribution, as shown in Figure \ref{figure structure2}. The process is formally defined as follows:
\begin{equation}
    \begin{array}{cc}
    \hspace{-0.36cm}
        p_{\theta}(X^{rec}_{0:K}{\mid}X^{obs}_{k} ) =  p(X^{rec}_{k}) \prod_{k=1}^{K} p_{\theta}(X^{rec}_{k-1}{\mid}X^{rec}_{k}, X^{rec}_{k} ),
    \end{array}
\end{equation}
Given the estimated distribution \( p(X^{rec}_{k}) \) at step \( k \) and the observation \( X^{obs}_{k} \), the diffusion network with parameter \( \theta \) predicts the distribution \( p(X^{rec}_{k-1}) \) at step \( k-1 \):
\begin{equation}
    \begin{aligned}
    &p_{\theta}(X^{rec}_{k-1} \mid X^{rec}_{k}, X^{obs}) = \\
    &\quad \quad \quad \quad \mathcal{N}(X^{rec}_{k-1}; \mu_{\theta}(X^{rec}_{k}, k, X^{obs}), \Sigma_{\theta}(X^{rec}_{k}, k)),
    \end{aligned}
    \label{eq7}
\end{equation}
where $ \mu_{\theta} $ is the estimated mean, and $\Sigma_{\theta}$ is the predicted covariance of the restored distribution of $ p(X^{rec}_{k-1}) $.

\subsection{Problem Definition}
Pedestrian intention prediction aims to forecast whether a target pedestrian will cross the street in a given scene by utilizing historical data. This data may include bounding boxes, which implicitly capture the pedestrian's intent, and the ego vehicle's speed, a critical factor influencing pedestrian behavior. This study proposes a diffusion-based model that integrates these two perceptual modalities to predict pedestrian crossing intentions, particularly in occlusion scenarios.

The historical bounding box of the observed pedestrians in two-dimensional Cartesian coordinates over the observed temporal period is given as $B = \left\{b_1, b_2, b_3, \dots, b_T \right\} \in \mathbb{R}^{T \times D_{b}} $, where $b_t = \left\{x^{TL}_{t}, y^{TL}_{t}, x^{BR}_{t}, y^{BR}_{t} \right\}$, with TL and BR denoting the top-left and bottom-right coordinates of the bounding box, respectively. Meanwhile, the center of the bounding box is denoted as $C = \left\{c_1, c_2, c_3, \dots, c_T \right\} \in \mathbb{R}^{T \times D_{c}} $, where $c_t = \left\{x^{C}_{t}, y^{C}_{t} \right\}$. The speed of the ego vehicle is defined as $V = \left\{v_1, v_2, v_3, \dots, v_T \right\} \in \mathbb{R}^{T \times D_{v}}$. The observed time horizon consists of $T$ frames indexed by $ t \in \left\{1, 2, \dots, T \right\} $. To simulate occlusion, $m$ frames are randomly selected from this sequence and fully masked, meaning that all modalities at those time steps are treated as unobservable.

The ground truth label $Y$ in the future is denoted as a binary label. The purpose of this work is to accurately estimate the intention $ \hat{Y} $ of the pedestrian based on the available observable historical motion features. Thus, pedestrian intention prediction with occluded observations involves training a model $ f(\cdot) $ with learnable parameters $\xi$ to estimate the intention $ \hat{Y} $ such that it closely approximates the true ground truth label $ Y $.
\begin{equation}
    \begin{array}{cc}
        \hat{Y} =  f( B, C, V, \xi ).
    \end{array}
\end{equation}
\begin{figure}[!ht]
\centering
	\includegraphics[width=8.5cm]{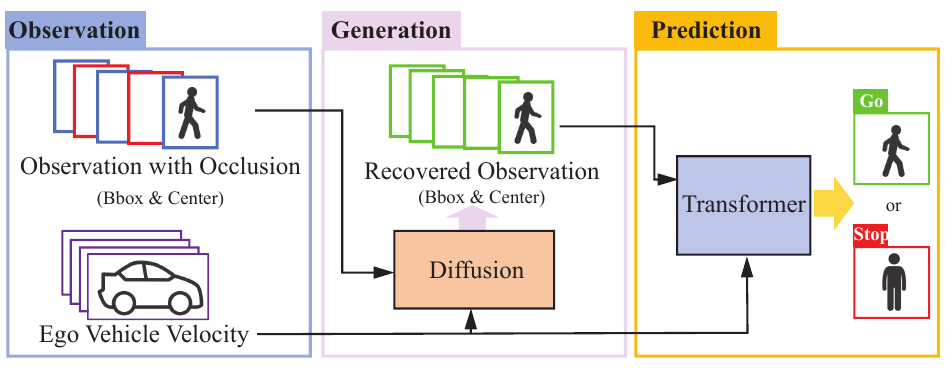}
	\caption{The overall framework of the ODM. The occluded observations are first embedded into the diffusion block to recover the missing motion features caused by occlusion. These recovered features are then used to estimate the crossing intention through the transformer block.
}
    \label{structure_1}
    \vspace{-1em}
\end{figure}
\begin{figure}[!ht]
    \centering{\includegraphics[width=8.8cm]{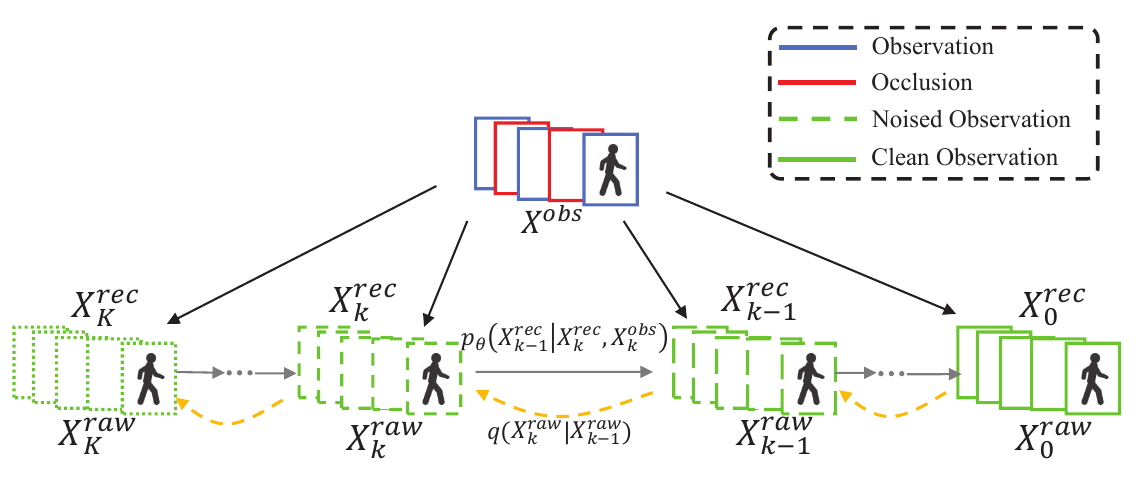}}
    \caption{Illustration of the diffusion process for motion indeterminacy variation. In the forward process, noise is gradually added to the raw observation sequences $X_{k}^{raw}$. In the reverse process, the added noise is removed by leveraging the clues provided by the occluded observations to recover observation $X_{k}^{rec}$.
 }
\label{diff_process} 
\vspace{-2em}
\end{figure}
\begin{figure}[!ht]
    \centering{\includegraphics[width=7cm]{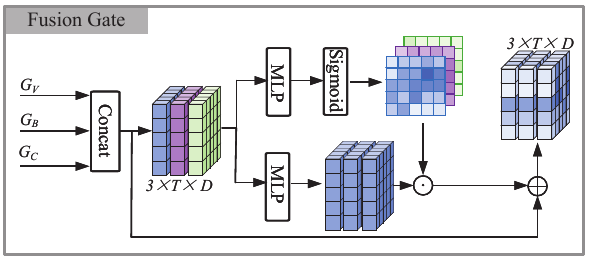}}
    \caption{Illustration of the fusion gate mechanism for integrating multimodal input features.
 }
\label{fussion_gate} 
\vspace{-1em}
\end{figure}

\subsection{Approach Overview}
The overall pipeline of ODM is shown in Figure \ref{structure_1}. The sequence of bounding boxes, combined with ego vehicle velocity, is processed by the diffusion model to reconstruct occluded motion from learned patterns. The diffusion process with occluded observations is illustrated in Figure \ref{diff_process}.

Specifically, in the forward process, observed traffic features are progressively corrupted with noise until they approximate a standard normal distribution. In the reverse process, features from occluded observations are fused with samples from the distribution for denoising, then processed by the proposed masked denoising network, as illustrated in Figure \ref{figure structure2}.

\begin{figure*}[!ht]\centering
	\includegraphics[scale= 0.65]{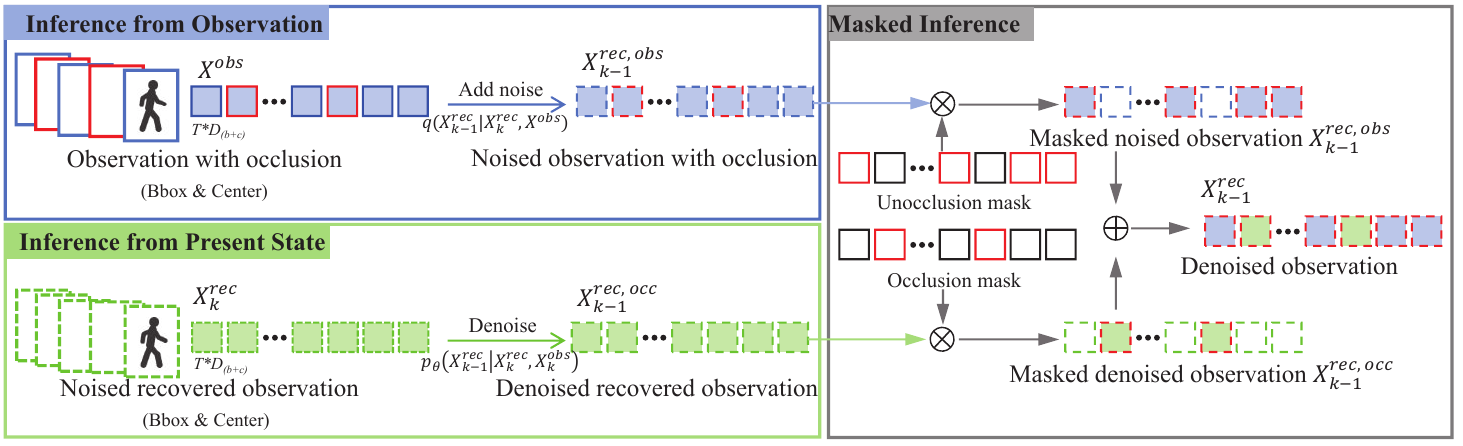}
	\caption{Illustration of the reverse mask mechanism. The recovered feature vector of $X^{rec}_{k-1}$ originates from two sources. For the observed parts, the features are directly calculated by adding noise to the observation. For the occluded parts, the features are denoised by the network. These two sources of features are then combined using the occlusion mask to create $X^{rec}_{k-1}$.
   }
    \label{diff_mask}
    \vspace{-1em}
\end{figure*}

In the reverse stage, the occlusion mask guides the diffusion model to generate features by effectively leveraging both observed data and occlusion patterns, as shown in Figure \ref{diff_mask}. The recovered motion observations are then fed into the transformer block to estimate pedestrian crossing intentions.

\subsection{Occlusion Masked Diffusion} 
In the inference stage, an occlusion-guided masking approach is introduced to generate features, leveraging the context of occlusion scenarios to constrain the sampling process at each reverse step. This enhances the model's ability to effectively utilize observable sequences and minimizes prediction errors during sequence generation, as illustrated in Figure \ref{diff_mask}.

Specifically, a binary mask matrix \(M= \{ m_i\}_{i=1}^{T} \in \mathbb{R}^{T \times D}\) with the same dimensions as the data vector \(X^{rec}_k\) is constructed, where \(m_i=1\) denotes masked tokens due to occlusion, while \(m_i=0\) indicates unmasked, observable tokens. The inference process starts by sampling \(X^{rec}_k\) from a standard Gaussian distribution \(\mathcal{N}(0, I)\). It is important to note that the vector \(X^{rec}_k\) consists of two components: the observed sequences \(X^{rec, obs}_k\) and the occluded sequences \(X^{rec, occ}_k\). For the observable components, the distribution of \(X^{rec, obs}_{k-1}\) can be directly computed using the posterior distribution \(q(X^{rec,obs}_{k-1} \mid X^{rec}_k, X^{obs})\), since both \(X^{obs}\) and \(X^{rec}_0\) share the same observed parts. Therefore, according to the forward posterior equation \eqref{eq3}, the distribution of observable sequences can be effectively formulated as:
\begin{equation}
      X^{rec,obs}_{k-1} \sim \mathcal{N}(X^{rec,obs}_{k-1}; \tilde{\mu}(X^{rec}_k, X^{obs}), \tilde{\beta}_{k}I), 
\end{equation}
where the mean $\tilde{\mu}(X^{rec}_k, X^{obs})$ and the variance $\tilde{\beta}_{k}$ matrix are given:
\begin{equation}
    \hspace{-0.24cm}
    \begin{array}{cc}
        \tilde{\mu}(X^{rec}_k, X^{obs}) = \frac{\sqrt{\bar{\alpha}_{k-1}}{\beta_{k}}}{1 - \bar{\alpha}_{k}}{X^{obs}} + \frac{\sqrt{\alpha_{k}}{(1-\tilde{\alpha}_{k-1})}}{1 - \bar{\alpha}_{k}}{X^{rec}_{k}},
    \end{array}
\end{equation}
\begin{equation}
    \begin{array}{cc}
        \tilde{\beta}_{k} = \frac{1 - \bar{\alpha}_{k-1}}{1 - \bar{\alpha}_{k}}{\beta_{k}}.
    \end{array}
\end{equation}

For the occluded parts, adapted from equation \eqref{eq7} they are specifically estimated using the learned parameterized distribution $p_{\theta}(X^{rec,occ}_{k-1} \mid X^{rec}_k, X^{obs}_{k})$, which is given as follows:
\begin{equation}
    X^{rec,occ}_{k-1} \sim \mathcal{N}(X^{rec,occ}_{k-1}; {\mu}_{\theta}(X^{rec}_k, X^{obs}_{k}, k), \tilde{\beta}_{k}I), 
\end{equation}
\begin{equation}
    \begin{aligned}
    &{\mu}_{\theta}(X^{rec}_k,k, X^{obs}_{k}) =  \\
    & \quad \quad \quad \frac{1}{\sqrt{\alpha_k}} (X^{rec}_k - \frac{\beta_k}{\sqrt{1-\bar{\alpha}_t}} {\epsilon}_{\theta}(X^{rec}_k, k, X^{obs}_{k})), 
    \end{aligned}
\end{equation}
where ${\epsilon}_{\theta}$ denotes the noise estimation neural network.

Consequently, the conditional distribution of $X^{rec}_{k-1}$ given $(X^{rec}_k, X^{obs}, M)$ is introduced as a mask-wise composition of two Gaussian distributions, incorporating occlusion masks into the reverse process:
\begin{equation}
\begin{aligned}
p_{\theta}\left(X^{rec}_{k-1} \mid X^{rec}_k, X^{obs}\right)& = \mathcal{N}\Big(X^{rec}_{k-1};  M \!\odot\! \mu_{\theta}(X^{rec}_k,k,X^{obs}_{k}) \\
 &  + (1\!-\!M) \!\odot\! \tilde{\mu}(X^{rec}_k,X^{obs}),\!\tilde{\beta}_{k} I \Big).
\end{aligned}
\end{equation}

Equivalently, sampling can be implemented as $X^{rec}_{k-1} = M \odot \tilde{X}^{occ}_{k-1} + (1-M) \odot \tilde{X}^{obs}_{k-1}$, where $\tilde{X}^{occ}_{k-1}$ and $\tilde{X}^{obs}_{k-1}$ are sampled results respectively. This formulation ensures that observable parts follow the posterior distribution, while occluded parts are inferred by the network, leading to more accurate reconstructions and preserving semantic consistency across the sequence.

\subsection{Occlusion-Aware Transformer for Noise Estimation}
The neural network employed in diffusion models is designed to predict noise $\epsilon_{\theta}$ at each reverse step. U-Net architectures were initially chosen for diffusion models due to their success in handling pixel-level autoregressive tasks \cite{ref81, ref82}. However, their effectiveness in processing sequential features, such as historical bounding box positions and vehicle speed data, especially in occlusion scenarios, requires further investigation. In this study, an occlusion-guided noise estimation network that enhances contextual representation with the mask modelling scheme in occlusion scenarios is proposed to handle sequential information, serving as a replacement for the U-Net. The module includes a feature extraction block and a masked transformer block as shown in the Figure. \ref{figure structure2}.

\subsubsection{Input Feature Extraction}Before feeding the observation features into the model for denoising, the input multimodal sequences undergo extraction, encoding, and fusion. In this stage, a simple yet effective modified gating-based multimodal feature re-weighting and aggregation mechanism \cite{ref93} is utilized to enhance feature representation.

Considering the low dimensionality of historical observation information with occlusion, including bounding boxes $B \in \mathbb{R}^{T \times D_{b}} $, $C \in \mathbb{R}^{T \times D_{c}} $, and vehicle speed $V \in \mathbb{R}^{T \times D_{v}} $, the input features are initially processed by fully connected layers to project them individually into a higher-dimensional space. Before extracting temporal dependencies by encoders, $\text{E}_{i}$, the sinusoidal positional encodings PE are added to encode the sequence order of the input. In this work, MLP layers contain double linear layers with a ReLU activation function.
\begin{equation}
\begin{aligned}
              G_{B} =  &\text{E}_{B}( \text{MLP}_{B}(B) + PE_{B}) \\
              G_{C} =  &\text{E}_{C}( \text{MLP}_{C}(C) + PE_{C}) \\
              G_{V} =  &\text{E}_{V}( \text{MLP}_{V}(V) + PE_{V}),
\end{aligned}
\end{equation}
where $G_{B} \in \mathbb{R}^{T \times D}$, $G_{C} \in \mathbb{R}^{T \times D}$, and $G_{V} \in \mathbb{R}^{T \times D}$. $\text{E}_{i}$ is a GRU-based motion encoder proposed to extract observed motion features, helping the model capture distant information and highlight important time steps in long sequences. Inside $\text{E}_{i}$, temporal context information is captured to model the evolution of pedestrian motion over time. It helps the model understand motion patterns and dependencies, particularly when occlusion causes missing motion features:
\begin{equation}
\alpha_{t,j} = \frac{\exp(v^{T} \tanh(W_{h} h_{t} + W_{x} h_{j}))}{\sum^{T}_{j=1} \exp(v^{T} \tanh(W_{h} h_{t} + W_{x} h_{j}))},
\end{equation}
\begin{equation}
c_t = \sum^{T}_{i=1} \alpha_{t,i} h_{i},
\end{equation}
where $h_t$ is the hidden state of the GRU at time $t$, and $W_h$ and $W_x$ are learnable parameters. Then, the updated hidden states with context information are given as:
\begin{equation}
    g_t = \text{FC}(\text{tanh}(W_{c}[c_t;h_t])),
\end{equation}
where $\text{FC}$ denotes a fully connected layer, and $W_c$ represents the learnable parameters.

Additionally, rather than directly fusing these features through concatenation, a gating mechanism is employed to adaptively select relevant features from each input modality, as illustrated in Figure \ref{fussion_gate}. Compared with static fusion approaches such as concatenation or averaging, the proposed gating mechanism adaptively emphasizes informative modalities while suppressing less reliable ones, which is beneficial under occlusion scenarios where certain inputs may become incomplete or noisy. The contribution of each feature is weighted by learnable parameters, which determine the influence of each modality within the entire input. The process is given as:
\begin{equation}
\begin{aligned}
            &H = \text{Sigmoid}( \text{MLP}_1(G) ) \odot \text{MLP}_2(G) + G,
\end{aligned}
\end{equation}
where $\text{Sigmoid}$ is the Sigmoid function, $G=[G_{B}, G_{C}, G_{V}]$. Together with the diffusion step $k$, the feature vector of observation is represented as:
\begin{equation}
\begin{aligned}
        X^{obs}_{k} = H + \text{MLP}_{K}( \text{Pos.emb.}(k) ), 
\end{aligned}
\end{equation}
where Pos.emb. indicates positional embedding layer, $X^{obs}_{k} \in \mathbb{R}^{T \times D}$.
 
\subsubsection{Occlusion-Masked Transformer} To enhance contextual reasoning and improve relational learning among semantic parts in occlusion scenarios, an occlusion masking block is introduced, which explicitly integrates occlusion cues into feature learning. Moreover, the occlusion-guided spatial-temporal encoder and decoder are designed to adaptively shift the focus on learning motion dynamics in response to occlusion patterns.

(1) Occlusion-Aware Spatial-Temporal Encoder and Decoder: Since ego-vehicle observations capture visible sequences with occluded contexts, they are crucial for modeling the correlation between occlusion patterns and pedestrian motion dynamics. To address this, we introduce occlusion-aware spatial-temporal motion reasoning encoder and decoder blocks, which enhance the prediction of noise features. 

The extracted observation feature $X^{obs}_{k}$, containing rich occlusion semantic information, is used to infer motion dynamics within the noisy motion sequences $H^{raw}_{k,in}$. Inspired by \cite{ref58}, we introduce a modified deformable attention mechanism with adaptive layer normalization. Instead of appending conditional tokens $X^{obs}_{k}$ to the input sequence $H^{raw}_{k,in}$, we propose a novel approach that incorporates $X^{obs}_{k}$ by regressing scale $Sc_{i}$, shift $Sh_{i}$, and offset $X_{k, i}$ parameters. This modification dynamically adapts feature flow, enabling more accurate reasoning in complex occlusion scenarios.

From a temporal perspective, the block is specifically designed to learn temporal offsets associated with occluded observation features, guiding the model to capture relationships between occlusion patterns and observable dynamics over time. Additionally, spatial offsets correlated with occlusion patterns are also learned, enabling the model to reason about motion at each frame in a more context-aware manner.

In detail, the trainable position $P_{k,T} \in \mathbb{R}^{T}$, associated with the occlusion semantic feature $X_{k,T}$, is distributed over the temporal dimensions of $H^{raw}_{k,Tscn}$, obtained by scaling and shifting the normalized $H^{raw}_{k,in}$. The initial position comes from a 1D uniform grid of length $T$. Then, the temporal offsets learned by the offset network adjust $P_{k,T}$ to obtain the semantic occlusion feature $H^{raw}_{k,Tsem}$, sampled from $H^{raw}_{k,Tscn}$. This process is defined as:
\begin{equation}
    \hspace{-0.05cm}
    Sc_1, Sh_1, Sc_2, X_{k,T}, Sc_3, Sh_2, Sc_4,X_{k,D},  =  \text{MLP}_{Ada} (X^{obs}_{k}),
\end{equation}
\begin{equation}
    H^{raw}_{k,Tscn} = \text{Norm}(H^{raw}_{k,in}) \odot Sc_1 + Sh_1, 
\end{equation}
\begin{equation}
    H^{raw}_{k,Tsem} = \phi ( H^{raw}_{k,Tscn}, \psi(P_{k,T},\theta_{Toff}(X_{k,T})) ),
\end{equation}
where $\text{Norm}$ denotes normalization, $\theta_{Toff}$ is the temporal offset network, $\psi$ represents the offset operation, and $\phi$ indicates the sampling operation using the linear interpolation method.

Subsequently, $H^{raw}_{k,Tsem}$ is mapped into key embeddings $K_{k,Tsem}$ and value embeddings $V_{k,Tsem}$ via separate linear projections. Query embeddings $Q_{k,T}$ are derived from a linear projection of $H^{raw}_{k,Tscn}$. The aggregated temporal semantic features $H^{raw}_{k,T}$ are then obtained through scaled multi-head attention with a residual connection:
\begin{equation}
    H^{raw}_{k,T} =  \text{MultiAtt}(Q_{k,T}, K_{k,Tsem}, V_{k,Tsem} )\odot Sc_2 + H^{raw}_{k,in},
\end{equation}
where $\text{MultiAtt}$ refers to the multi-head attention.

Similarly, in the subsequent stage, the above process is applied to the spatial dimension of $H^{raw}_{k,T}$, which can be described as:
\begin{equation}
    H^{raw}_{k,Dscn} = \text{Norm}(H^{raw}_{k,T}) \odot Sc_3 + Sh_2, 
\end{equation}
\begin{equation}
    H^{raw}_{k,Dsem} = \phi ( H^{raw}_{k,Dscn}, \psi(P_{k,D},\theta_{Doff}(X_{k,D})) ),
\end{equation}
\begin{equation}
    \hspace{-0.05cm}
    H^{raw}_{k,out} =  \text{MultiAtt}(Q_{k,D}, K_{k,Dsem}, V_{k,Dsem} )\odot Sc_4 + H^{raw}_{k,Tscn},
\end{equation}
where $\theta_{Doff}$ represents the spatial offset network, $P_{k,D} \in \mathbb{R}^{D}$ is the learnable spatial position, $K_{k,Dsem}$ and $V_{k,Dsem}$ are derived from $H^{raw}_{k,Dsem}$ via separate linear projections, and $Q_{k,D}$ is derived from a linear projection of $H^{raw}_{k,Dscn}$.

It is worth noting that during the inference stage, $H^{raw}_{k,in}$ is replaced by $ H^{rec}_{k,in} $, which is the feature sampled from the previous step $ k+1$ according to equation \eqref{eq7}. Benefiting from the proposed occlusion-aware motion reasoning process, the correlation between occlusion patterns and motion dynamics is effectively explored, resulting in the extraction of high-quality features that enhance the performance of noise estimation.

(2) {Occlusion Masking Block}: In this block, the motion features from the encoder output, $ Enc_{out} \in \mathbb{R}^{T \times D} $, are processed through a network that mirrors the encoder-decoder structure to predict the occlusion-masked embedding. Next, these two features, the input and predicted features, are combined using a simple yet effective method, a shortcut connection with the previously defined occlusion masks. The process is given as:
\begin{equation}
    {Dec}_{in} =   \text{Trans}({Enc}_{out}) \odot M + ({Enc}_{out}) \odot (1-M),
\end{equation}
where $\text{Trans}$ is the transformer block, $M$ is occlusion mask.

The block predicts the tokens masked corresponding to occlusion patterns, while the unmasked tokens, representing observable parts, are retained. This latent space approach explicitly enhances the model's ability to learn occluded features from a semantic perspective. Meanwhile, it prevents the model from overemphasizing the reconstruction of occluded regions at the expense of the diffusion process.
\vspace{-0.3em}

\subsection{Intention Prediction Block}
After the observation sequences are generated through the diffusion process, an intention prediction block, implemented as a transformer, is used to estimate pedestrian intentions. The structure mirrors that of a standard transformer, comprising a multi-head self-attention layer and a feedforward layer with residual connections. A softmax activation is then applied to predict intent assignment probabilities.

\subsection{Training Optimization}
The proposed framework follows a multi-task learning paradigm, jointly optimizing both an occlusion reconstruction task and an intention classification task. The reconstruction task ensures that occluded observations are faithfully recovered, while the classification task focuses on the ultimate goal of accurately predicting pedestrian crossing intentions.

For the diffusion model, the primary training objective is to maximize the adapted Variational Lower Bound (VLB) \cite{ref76} as directly optimizing $\mathbb{E}[\log p_{\theta}(X^{rec}_0)]$ is difficult. This reconstruction loss effectively encourages the diffusion network to recover occluded motion features, which then serve as refined inputs for the downstream intention classification task.
\begin{equation}
    \begin{aligned}
        & \mathbb{E}[\log p_{\theta}(X^{rec}_0)] \geq \mathbb{E}_q [ \log \frac{p_{\theta}(X^{rec}_{0:K} | X^{obs})}{q(X^{raw}_{1:K} | X^{raw}_0)} ] \\
        & = \mathbb{E}_q [ \log p(X^{rec}_K) + \sum_{k=1}^{K} \log \frac{p_{\theta}(X^{rec}_{k-1} | X^{rec}_k, X^{obs}_{k})}{q(X^{raw}_k | X^{raw}_{k-1})} ].
    \end{aligned}
\end{equation}

This process is simplified by using the denoising network to estimate the noise at each diffusion step. The objective is formulated as the mean squared error (MSE) between the predicted and true noise, as shown below:
\begin{equation} 
 {\mathcal L}_{simp}(\theta) =  \mathbb{E}_{\epsilon, X^{raw}_0, k} \parallel \epsilon - {\epsilon}_\theta(X^{raw}_k, k, X^{obs}_{k}) \parallel ^2,
\end{equation}
where ${\epsilon}$ represents the ground truth noise at step $k$ during the forward process, and ${\epsilon}_\theta$ denotes the denoising network.

For intention prediction, the task is framed as a binary classification problem that estimates the probability of a pedestrian intending to cross or not. In this context, the cross-entropy loss function is employed. The total loss is then formulated as a weighted combination of the diffusion model loss and the intention prediction model loss, expressed as:
\begin{equation}
    {\mathcal L}^{i}_{int}=  - y^{i} \cdot \text{log}(\hat{y}^{i}) - (1-y^{i}) \cdot \text{log}(1-\hat{y}^{i}),
\end{equation}
\begin{equation}
    {\mathcal L}= \frac{1}{N} \sum_{i=1}^{N}  ({\mathcal L}^{i}_{simp}+ \lambda {\mathcal L}^{i}_{int}),
\end{equation}
where $\lambda$ is an empirical hyperparameter fixed at 1.2 to balance the auxiliary reconstruction and primary classification tasks, ensuring both objectives contribute during training, $N$ is the total number of samples, $\hat{y}^i$ indicates the predicted intention, and $y^i$ corresponds to the ground-truth intention. 

\section{Experiments and DISCUSSIONS}
To comprehensively assess the performance of the proposed ODM, we evaluate the model on two public benchmark datasets, namely PIE and JAAD. Extensive experiments are conducted under diverse occlusion scenarios to better simulate real-world conditions. The results are thoroughly analyzed and compared with state-of-the-art methods, including \cite{ref90, ref74}, both quantitatively and qualitatively. Detailed ablation studies are also performed with different model configurations to validate the effect and functionality of each component.

\subsection{Dataset}
Our experiment is conducted on two intention prediction benchmark datasets: the JAAD (Joint Attention for Autonomous Driving) dataset \cite{ref63} and the PIE (Pedestrian Intention Estimation) dataset \cite{ref64}. The JAAD dataset consists of recorded on-board video clips, containing 346 short sequences extracted from over 240 hours of driving footage. In contrast, the PIE dataset captures continuous traffic flow in urban street scenes, offering a comprehensive representation of real-world dynamics. Additionally, vehicle information, including speed, heading direction, and GPS coordinates, synchronized with the video footage, is obtained from onboard diagnostics sensors.

Both the JAAD and PIE datasets share a broadly similar annotation framework, where pedestrians are labeled with future activities indicating whether they will cross or not cross. This defines the task as a binary classification problem, where the model predicts crossing probability and is directly evaluated against the ground truth. A key difference between the two datasets lies in the representation of vehicle motion: in PIE, the ego-vehicle speed labels are explicitly provided and used as input, whereas in JAAD, only vehicle motion activity labels, such as accelerating and decelerating, are available, which we adopt as a practical proxy for vehicle velocity.

Following previous methods, we use observation sequences of 15 frames, including ego-vehicle speed and pedestrian bounding boxes, as inputs to estimate pedestrian crossing intentions at the 16th frame. To simulate occlusion scenarios, portions of the observation sequences are randomly masked with varying lengths (1 to 5 frames) and two masking patterns (EO and PO). For each occlusion length and pattern, we conduct separate experiments (EO1 to EO5, PO1 to PO5). 

\begin{figure}[!ht]\centering
	\includegraphics[width=8.8cm]{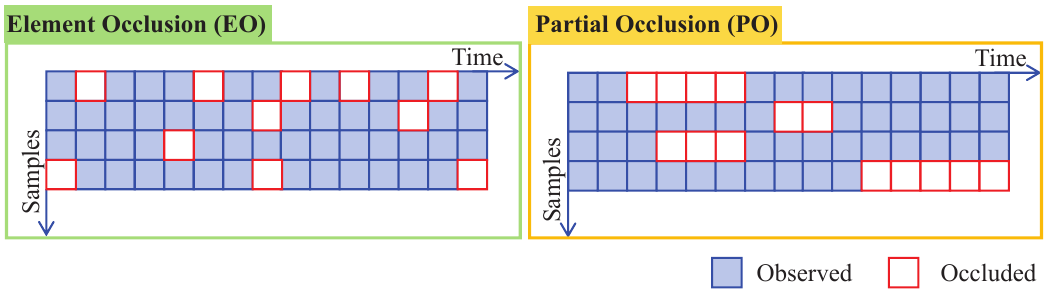}
	\caption{The left figure represents Element Occlusion (EO), where data is randomly occluded, while the right figure represents Partial Occlusion (PO), where data is continuously occluded.}
    \label{occlus_pattern}
\vspace{-1em}
\end{figure}

As illustrated in Figure \ref{occlus_pattern}, Element Occlusion (EO) refers to cases where several non-consecutive frames in a sequence are fully masked, such that a pedestrian observable at time $t-1$ becomes unobservable at $t$. Partial Occlusion (PO) refers to cases where multiple consecutive frames are masked, for instance, when a pedestrian visible at $t-1$ is missing at $t$, $t+1$, and $t+2$. In both scenarios, the occluded frames are chosen at random within the observation sequence, and the start positions in PO are also randomly sampled. In both EO and PO settings, the occlusions correspond to complete unobservability of the pedestrian information at selected time steps, and our model reconstructs the missing observations. For simplicity, the occlusion simulation does not explicitly depend on scene complexity or pedestrian position but instead assumes temporally random missing observations.
\subsection{Metrics}
To evaluate the performance of the proposed method on the intention prediction task, three widely used standard metrics commonly employed in previous studies are used: Accuracy (Acc), Area Under the Curve (AUC), and F1-score (F1).

(1) Accuracy is a widely used fundamental metric in classification tasks, mathematically defined as the proportion of correctly classified samples to the total number of samples.
\begin{equation}
         Accuracy=\frac{ TP + TN }{TP + TN + FP + FN},  
\end{equation}
where $TP$ denotes True Positive, $TN$ denotes True Negative, $FP$ denotes False Positive, and $FN$ denotes False Negative.

(2) Area Under the Curve (AUC) is a metric for binary classification that measures the ability to distinguish between positive and negative classes. A perfect model has an AUC of 1.0, while a random classifier has 0.5.

(3) F1-score is an especially important metric often used to rigorously assess the performance of classification models, particularly in real-world cases involving imbalanced datasets.
\begin{equation}
         F1=2 \times \frac{ Precision \times Recall }{Precision + Recall},  
\end{equation}
where $precision = TP / (TP + FP)$ measures the proportion of predicted positive cases that are actually positive, and $Recall = TP / (TP + FN)$ measures the proportion of actual positive cases that are correctly predicted.

\subsection{Implementation}
The model is implemented using the PyTorch framework and trained on a powerful environment fully equipped with Nvidia RTX 4090 GPUs and an Intel Xeon 4210R CPU. In this work, the embedding dimensions of all input modality tensors are projected to 64 to ensure dimensional consistency. For the masked transformer block, both the encoder and decoder layers are set to 2, and a single layer of the occlusion masking block is adopted. The attention head is set to 8. For the intention prediction block, the transformer layer is set to 4 with 4 heads. The diffusion step is set to 100. During training, we use the Adam optimizer with a fixed learning rate of $1 \times 10^{-4}$, a batch size of 64, and apply a dropout rate of 0.1.

\subsection{Baseline Methods}
In this work, several studies that demonstrate state-of-the-art performance on both the PIE and JAAD benchmarks are included as baseline models. All models are evaluated on the datasets PIE and JAAD with occluded observations.   

\textbf{MultiRNN}\cite{ref92}: Several RNN branches are used to process different feature sources, and the features are then concatenated into a fully connected layer for prediction.

\textbf{SFRNN}\cite{ref91}: The model is built on stacked RNNs, where diverse types of input features are carefully processed and effectively fused hierarchically at each separate level.

\textbf{I3D}\cite{ref57}: The network takes RGB frame stacks as input and generates the prediction through a fully connected layer.

\textbf{PCPA}\cite{ref90}: The model employs a dedicated convolutional branch to encode visual features and parallel RNN branches to process diverse non-image features. These features are then seamlessly integrated using a modality attention mechanism.

\textbf{TrEP}\cite{ref74}: A novel transformer-based model for robust and efficient uncertainty-aware pedestrian intention estimation using bounding boxes and ego-vehicle speed as input.

\subsection{Quantitative Evaluation}
In this section, we evaluate and compare the accuracy of predicted future intentions. The quantitative experimental results on the PIE and JAAD datasets under both EO and PO scenarios are presented in Tables \ref{Table 2} and \ref{Table 3}, respectively. In each case, the occlusion lengths vary from 1 to 5 frames.

It is worth noting that models with fewer input modalities still maintain relatively high performance under challenging occlusion scenarios. For instance, PCPA, I3D, SFRNN, and MultiRNN use diverse modality data, including bounding boxes, poses, images, and speed, as input, while the proposed model and TrEP only leverage bounding boxes and ego-vehicle speed for intention prediction. The underlying reason behind this trend may be that incorporating more modality inputs under occlusion leads to the loss of a larger set of features, which significantly impacts prediction performance.


Compared to other methods that do not explicitly account for missing motion patterns, the proposed method consistently improves all three metrics across all datasets and scenarios, suggesting the strong potential importance of considering occluded motion patterns. Specifically, in the EO5 scenario with the PIE dataset, the proposed method outperforms TrEP, achieving maximum improvements of 5\%, 4\%, and 5\% in Acc, AUC, and F1, respectively. For the JAAD dataset, the proposed method also clearly outperforms all the compared methods. For example, in the EO5 scenario, improvements of 4\%, 3\%, and 3\% in Acc, AUC, and F1, respectively, are observed. These consistent gains highlight the robustness of our framework in handling severe occlusion, and they further demonstrate its value for safety-critical applications such as autonomous driving and intelligent surveillance systems.

\begin{table*}[t]
\centering
\begin{minipage}[t]{0.48\textwidth}
    \centering
    \fontsize{7.5}{6}\selectfont
    \setlength{\tabcolsep}{2.5 pt}
    \renewcommand{\arraystretch}{1.0}
    \caption{Performance Compared to Proposed Method on PIE and JAAD Dataset in Element Occlusion (EO). 
    Input modalities: B = Bounding box, I = Image, P = Pose, V = Velocity, \textbf{Bold} is better.} 
    \vspace{-1em}
    \label{Table 2}
    \begin{tabular}{p{1.8cm} | p{0.8cm} | >{\centering\arraybackslash}m{0.8cm} | c c c | c c c}
     \toprule
    \multirow{2}{*}{Method} & \multirow{2}{*}{Inputs} & \multirow{2}{*}{Type} 
     & \multicolumn{3}{c|}{PIE} & \multicolumn{3}{c}{JAAD} \\ \cmidrule(lr){4-6} \cmidrule(lr){7-9}
     &  &  & Acc $\uparrow$ & AUC $\uparrow$ & F1 $\uparrow$ & Acc $\uparrow$ & AUC $\uparrow$ & F1 $\uparrow$ \\
    \midrule                                           
    MultiRNN \cite{ref92} & B,I & \multirow{6}{*}{EO,1} & 0.80 & 0.78 & 0.67 & 0.76 & 0.75 & 0.53 \\
    SFRNN\cite{ref91}     & B,I,P,V & & 0.82 & 0.77 & 0.67 & 0.84 & 0.74 & 0.41 \\ 
    I3D\cite{ref57}       & I & & 0.82 & 0.80 & 0.71 & 0.85 & 0.70 & 0.63 \\ 
    PCPA\cite{ref90}      & B,I,P,V & & 0.83 & 0.81 & 0.71 & 0.86 & 0.70 & 0.60 \\ 
    TrEP\cite{ref74}      & B,V & & 0.89 & 0.93 & 0.88 & 0.85 & 0.92 & 0.86 \\ 
    Ours & B,V & & \textbf{0.91} & \textbf{0.95} & \textbf{0.90} & \textbf{0.88} & \textbf{0.94} & \textbf{0.88} \\
    \midrule
    MultiRNN & B,I & \multirow{6}{*}{EO,2} & 0.77 & 0.76 & 0.65 & 0.75 & 0.75 & 0.52 \\
    SFRNN & B,I,P,V & & 0.80 & 0.76 & 0.65 & 0.84 & 0.73 & 0.40 \\
    I3D & I & & 0.78 & 0.77 & 0.66 & 0.82 & 0.69 & 0.61 \\ 
    PCPA & B,I,P,V & & 0.78 & 0.78 & 0.66 & 0.84 & 0.70 & 0.60 \\ 
    TrEP & B,V & & 0.88 & 0.93 & 0.88 & 0.85 & 0.91 & 0.85 \\ 
    Ours & B,V& & \textbf{0.91} & \textbf{0.95} & \textbf{0.90} & \textbf{0.88} & \textbf{0.94} & \textbf{0.87} \\
    \midrule
    MultiRNN & B,I & \multirow{6}{*}{EO,3} & 0.75 & 0.74 & 0.65 & 0.74 & 0.74 & 0.52 \\ 
    SFRNN & B,I,P,V & & 0.79 & 0.75 & 0.64 & 0.83 & 0.72 & 0.40 \\ 
    I3D & I & & 0.76 & 0.76 & 0.65 & 0.81 & 0.68 & 0.61 \\ 
    PCPA & B,I,P,V & & 0.73 & 0.76 & 0.63 & 0.82 & 0.71 & 0.59 \\ 
    TrEP & B,V & & 0.88 & 0.92 & 0.87 & 0.85 & 0.91 & 0.85 \\ 
    Ours & B,V& & \textbf{0.90} & \textbf{0.95} & \textbf{0.90} & \textbf{0.87} & \textbf{0.93} & \textbf{0.87} \\
    \midrule
    MultiRNN & B,I & \multirow{6}{*}{EO,4} & 0.71 & 0.70 & 0.64 & 0.72 & 0.74 & 0.50 \\ 
    SFRNN & B,I,P,V & & 0.78 & 0.74 & 0.63 & 0.82 & 0.72 & 0.41 \\ 
    I3D & I & & 0.75 & 0.74 & 0.64 & 0.78 & 0.66 & 0.60 \\ 
    PCPA & B,I,P,V & & 0.70 & 0.75 & 0.62 & 0.79 & 0.72 & 0.58 \\ 
    TrEP & B,V & & 0.86 & 0.92 & 0.86 & 0.84 & 0.91 & 0.84 \\ 
    Ours & B,V& & \textbf{0.90} & \textbf{0.95} & \textbf{0.89} & \textbf{0.87} & \textbf{0.94} & \textbf{0.86} \\
    \midrule
    MultiRNN & B,I & \multirow{6}{*}{EO,5} & 0.66 & 0.69 & 0.61 & 0.72 & 0.73 & 0.49 \\ 
    SFRNN & B,I,P,V & & 0.77 & 0.73 & 0.61 & 0.81 & 0.72 & 0.42 \\ 
    I3D & I & & 0.72 & 0.73 & 0.61 & 0.75 & 0.62 & 0.59 \\
    PCPA & B,I,P,V & & 0.70 & 0.73 & 0.60 & 0.73 & 0.72 & 0.57 \\ 
    TrEP & B,V & & 0.85 & 0.91 & 0.85 & 0.83 & 0.90 & 0.83 \\ 
    Ours & B,V& & \textbf{0.90} & \textbf{0.95} & \textbf{0.90} & \textbf{0.87} & \textbf{0.93} & \textbf{0.86} \\
    \bottomrule
    \end{tabular}
\end{minipage}
\hspace{0.02\textwidth}
\begin{minipage}[t]{0.48\textwidth}
    \centering
    \fontsize{7.5}{6}\selectfont
    \setlength{\tabcolsep}{2.5 pt}
    \renewcommand{\arraystretch}{1.0}
    \caption{Performance Compared to Proposed Method on PIE and JAAD Dataset in Partial Occlusion (PO). 
    Input modalities: B = Bounding box, I = Image, P = Pose, V = Velocity, \textbf{Bold} is better.} 
    \vspace{-1em}
    \label{Table 3}
    \begin{tabular}{p{1.6cm} | p{0.8cm} | >{\centering\arraybackslash}m{0.8cm} | c c c | c c c}
     \toprule
    \multirow{2}{*}{Method} & \multirow{2}{*}{Inputs} & \multirow{2}{*}{Type} 
     & \multicolumn{3}{c|}{PIE} & \multicolumn{3}{c}{JAAD} \\ \cmidrule(lr){4-6} \cmidrule(lr){7-9}
    &  &  & Acc $\uparrow$ & AUC $\uparrow$ & F1 $\uparrow$ & Acc $\uparrow$ & AUC $\uparrow$ & F1 $\uparrow$ \\
    \midrule  
    MultiRNN & B,I & \multirow{6}{*}{PO,1} & 0.80 & 0.77 & 0.68 & 0.78 & 0.75 & 0.53 \\ 
    SFRNN & B,I,P,V & & 0.82 & 0.77 & 0.67 & 0.84 & 0.73 & 0.41 \\ 
    I3D & I & & 0.81 & 0.80 & 0.71 & 0.84 & 0.70 & 0.62 \\ 
    PCPA & B,I,P,V & & 0.81 & 0.81 & 0.71 & 0.86 & 0.69 & 0.61 \\ 
    TrEP & B,V& & 0.88 & 0.92 & 0.87 & 0.86 & 0.91 & 0.85 \\ 
    Ours & B,V& & \textbf{0.91} & \textbf{0.94} & \textbf{0.90} & \textbf{0.87} & \textbf{0.94} & \textbf{0.87} \\
    \midrule  
    MultiRNN & B,I & \multirow{6}{*}{PO,2} & 0.80 & 0.76 & 0.67 & 0.76 & 0.75 & 0.51 \\ 
    SFRNN & B,I,P,V & & 0.80 & 0.76 & 0.65 & 0.83 & 0.72 & 0.40 \\ 
    I3D & I & & 0.78 & 0.76 & 0.66 & 0.81 & 0.68 & 0.61 \\ 
    PCPA & B,I,P,V & & 0.79 & 0.80 & 0.68 & 0.85 & 0.67 & 0.60 \\
    TrEP & B,V & & 0.87 & 0.92 & 0.87 & 0.85 & 0.91 & 0.86 \\ 
    Ours & B,V& & \textbf{0.90} & \textbf{0.94} & \textbf{0.89} & \textbf{0.87} & \textbf{0.94} & \textbf{0.86} \\
    \midrule  
    MultiRNN & B,I & \multirow{6}{*}{PO,3} & 0.79 & 0.76 & 0.65 & 0.74 & 0.74 & 0.50 \\ 
    SFRNN & B,I,P,V & & 0.78 & 0.74 & 0.63 & 0.82 & 0.71 & 0.40 \\ 
    I3D & I & & 0.76 & 0.72 & 0.64 & 0.79 & 0.65 & 0.60 \\ 
    PCPA & B,I,P,V & & 0.78 & 0.73 & 0.72 & 0.84 & 0.67 & 0.60 \\
    TrEP & B,V & & 0.87 & 0.91 & 0.87 & 0.84 & 0.92 & 0.85 \\ 
    Ours & B,V& & \textbf{0.90} & \textbf{0.94} & \textbf{0.90} & \textbf{0.86} & \textbf{0.93} & \textbf{0.86} \\
    \midrule  
    MultiRNN & B,I & \multirow{6}{*}{PO,4} & 0.78 & 0.74 & 0.65 & 0.73 & 0.73 & 0.50 \\ 
    SFRNN & B,I,P,V & & 0.77 & 0.73 & 0.61 & 0.82 & 0.71 & 0.41 \\
    I3D & I & & 0.75 & 0.71 & 0.62 & 0.78 & 0.64 & 0.60 \\ 
    PCPA & B,I,P,V & & 0.73 & 0.71 & 0.61 & 0.84 & 0.65 & 0.58 \\
    TrEP & B,V & & 0.86 & 0.91 & 0.86 & 0.84 & 0.91 & 0.84 \\ 
    Ours & B,V & & \textbf{0.90} & \textbf{0.94} & \textbf{0.89} & \textbf{0.86} & \textbf{0.94} & \textbf{0.85} \\
    \midrule  
    MultiRNN & B,I & \multirow{6}{*}{PO,5} & 0.76 & 0.74 & 0.64 & 0.72 & 0.73 & 0.48 \\ 
    SFRNN & B,I,P,V & & 0.75 & 0.72 & 0.60 & 0.81 & 0.70 & 0.40 \\
    I3D & I & & 0.72 & 0.70 & 0.60 & 0.74 & 0.62 & 0.59 \\
    PCPA & B,I,P,V & & 0.71 & 0.70 & 0.62 & 0.83 & 0.65 & 0.56 \\
    TrEP & B,V & & 0.85 & 0.91 & 0.86 & 0.83 & 0.91 & 0.83 \\ 
    Ours & B,V & & \textbf{0.89} & \textbf{0.93} & \textbf{0.89} & \textbf{0.86} & \textbf{0.93} & \textbf{0.86} \\
    \bottomrule
    \end{tabular}
\end{minipage}
\vspace{-1em}
\end{table*}

\subsection{Qualitative Evaluation}
\begin{figure*}[t]
    \centering
    \subfloat[][$k=100$]{\includegraphics[width=0.20\textwidth]{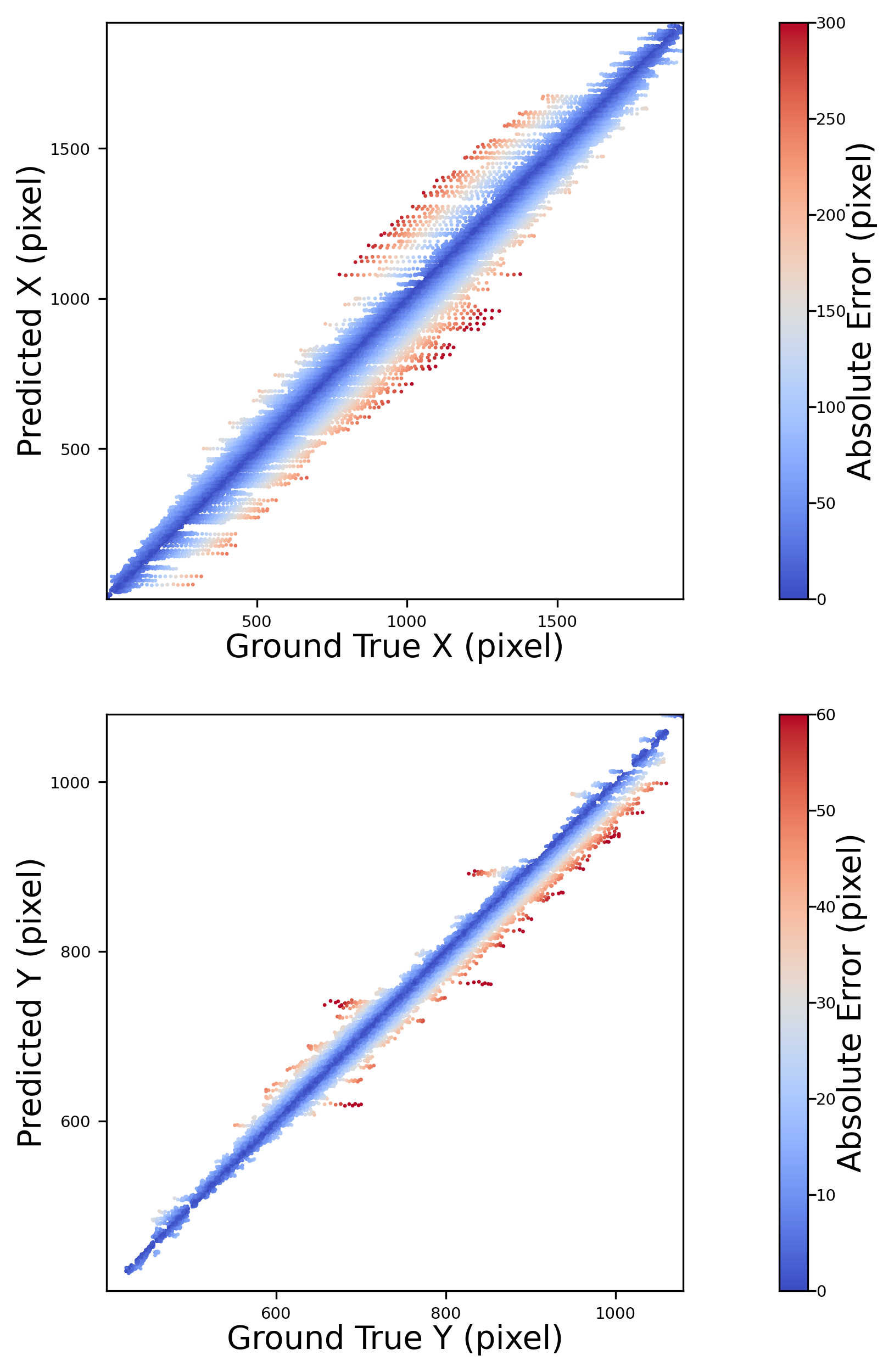}} 
     \hspace{-0.01\textwidth}
    \subfloat[][$k=75$]{\includegraphics[width=0.20\textwidth]{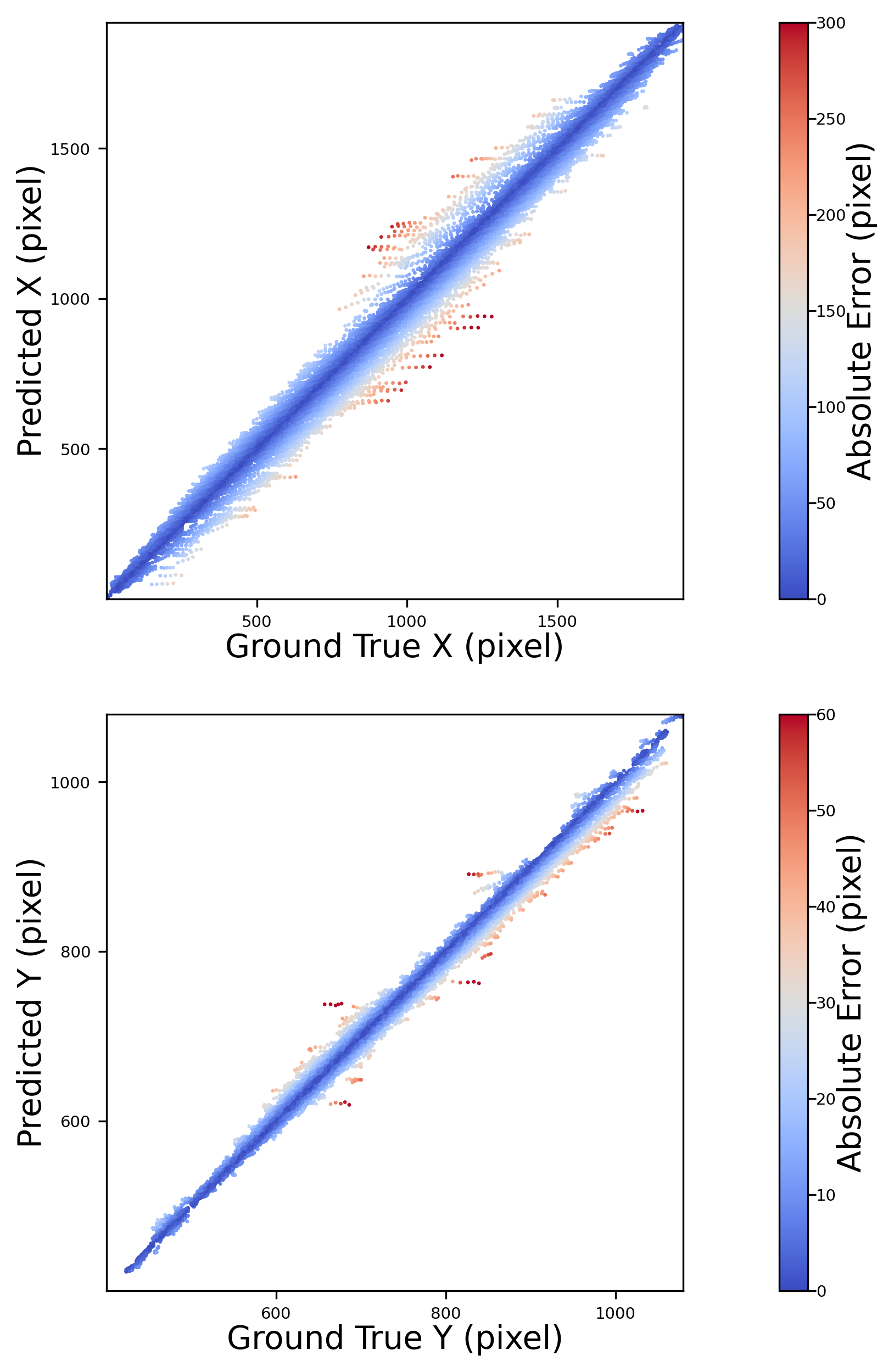}}
     \hspace{-0.01\textwidth}
    \subfloat[][$k=50$]{\includegraphics[width=0.20\textwidth]{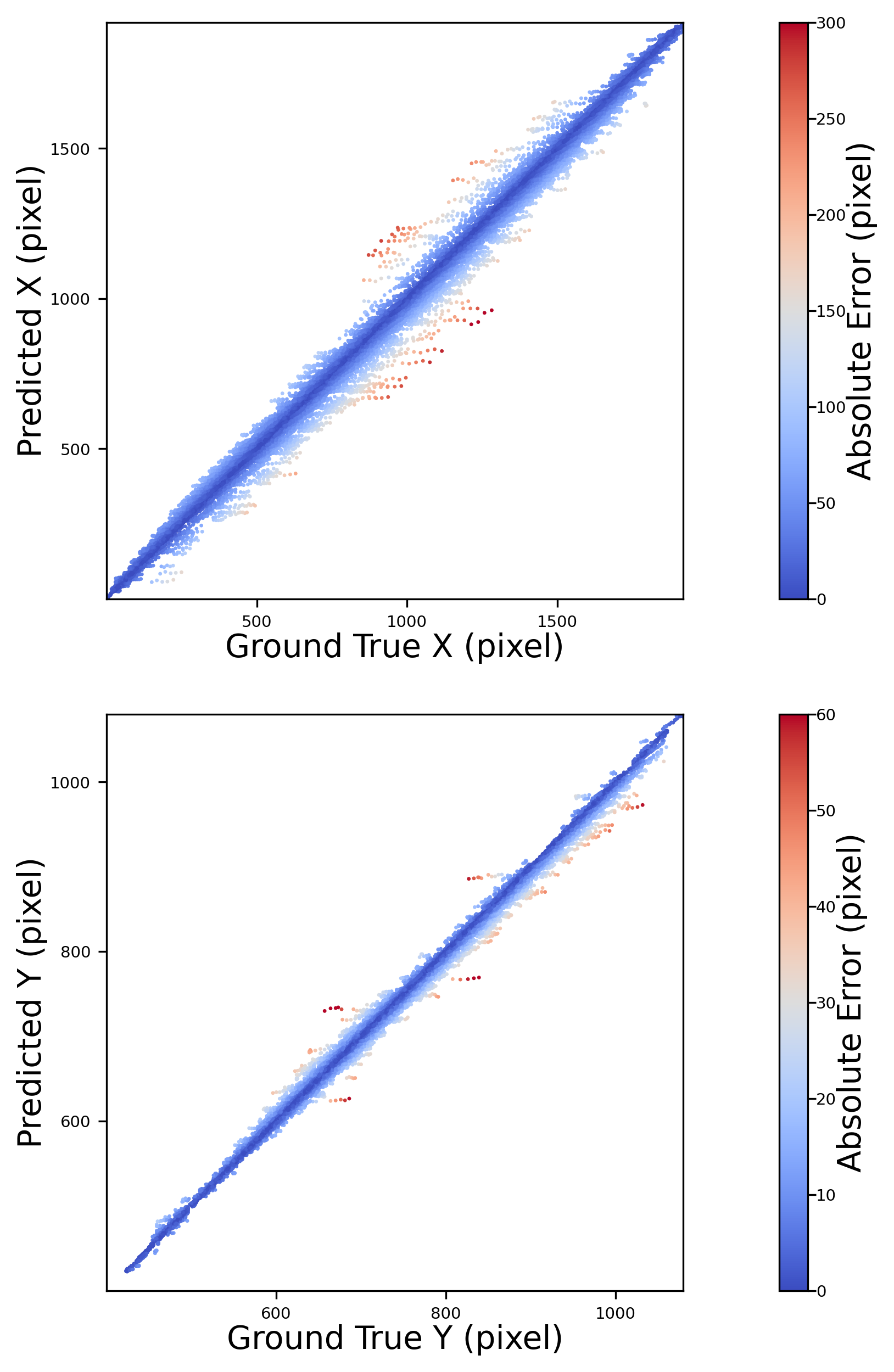}} 
     \hspace{-0.01\textwidth}
    \subfloat[][$k=25$]{\includegraphics[width=0.20\textwidth]{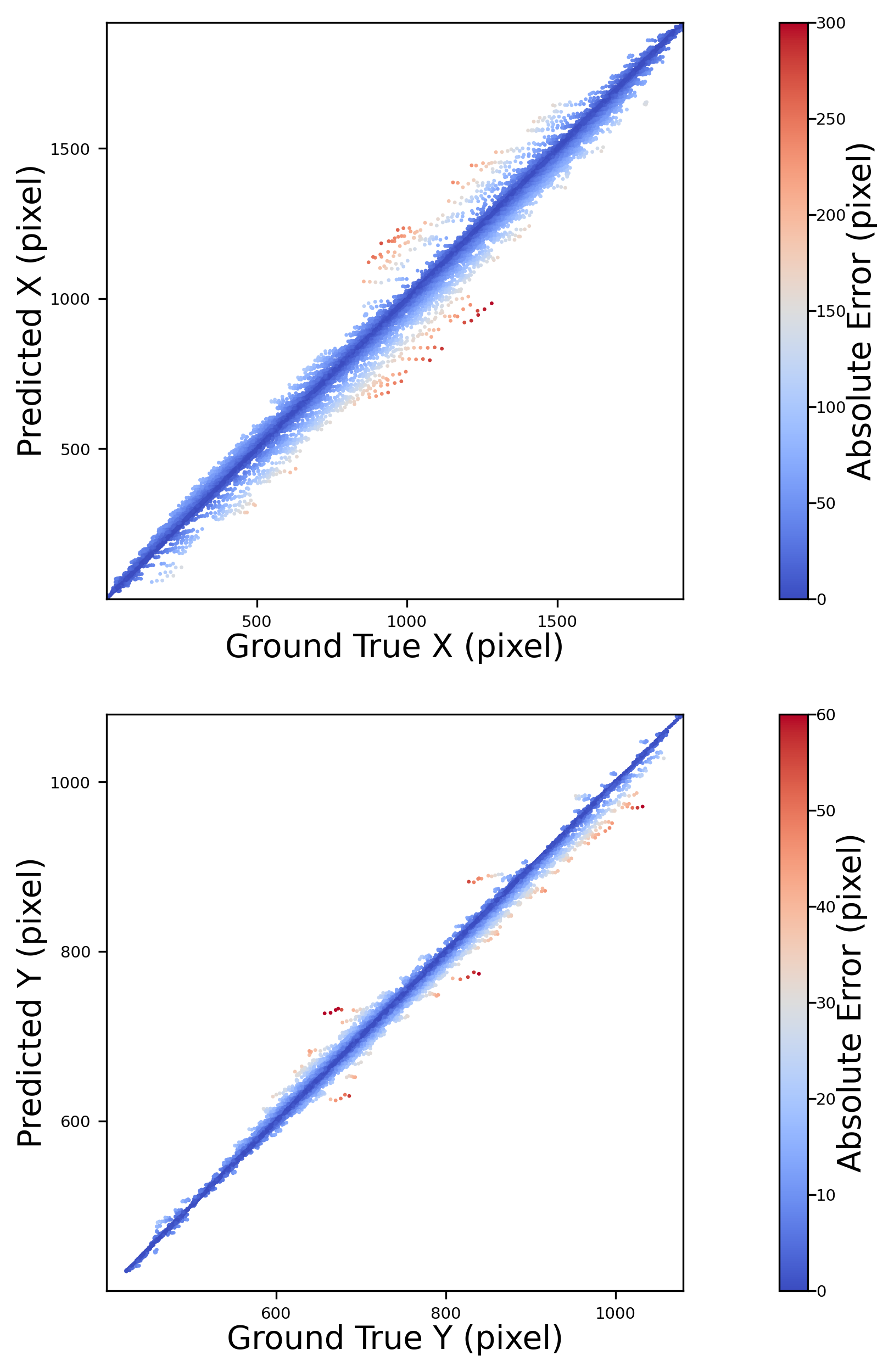}}
     \hspace{-0.01\textwidth}
    \subfloat[][$k=0$]{\includegraphics[width=0.20\textwidth]{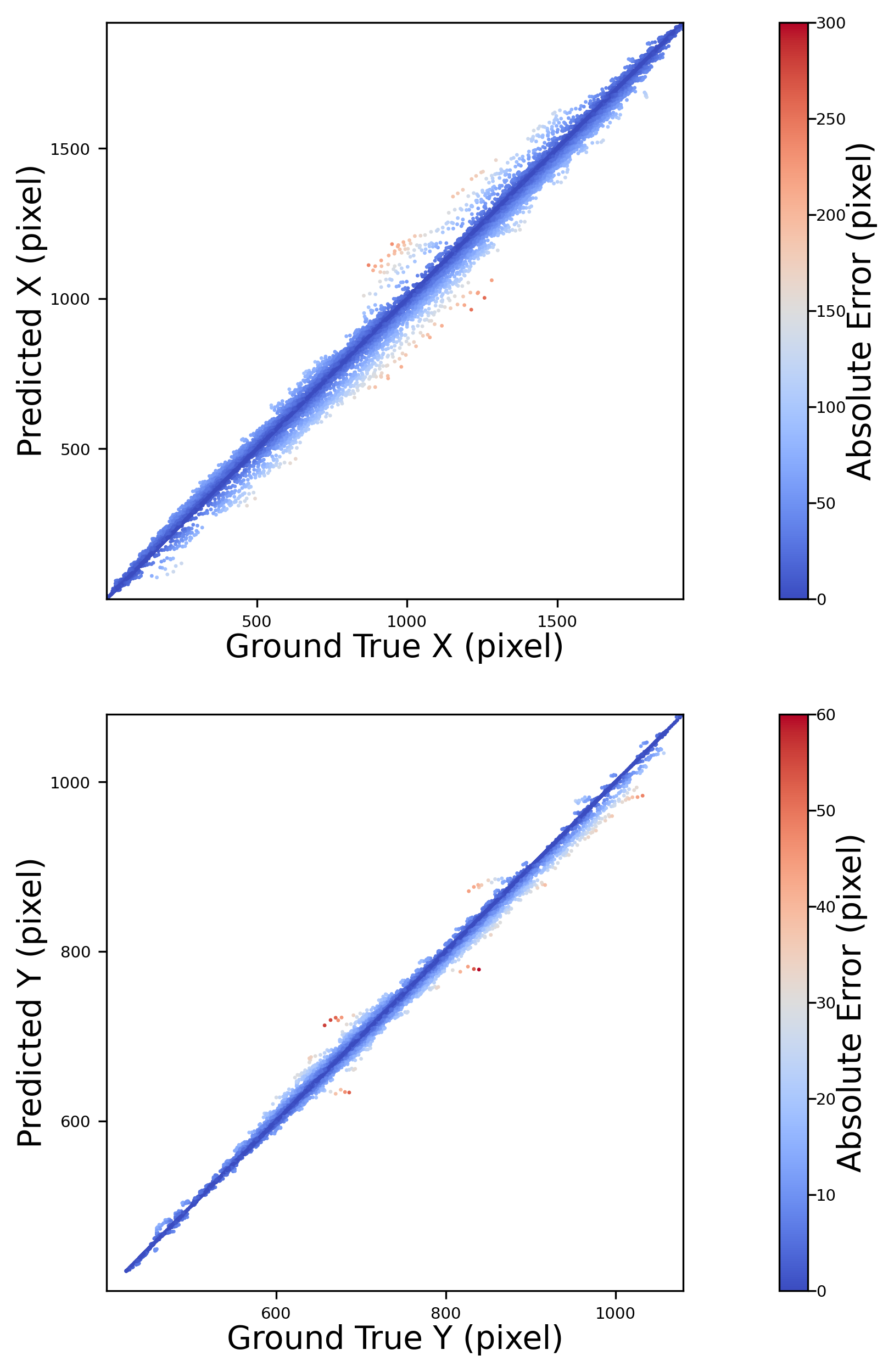}} 
    \caption{The visualization of the denoising process for the observations. The top row displays the denoising of the X coordinates of the bounding boxes, while the bottom row shows the denoising of the Y coordinates. In our study, the diffusion step $K$ is 100.}
    \label{Diff_ProcessV}
    \vspace{-1em}
\end{figure*}
\subsubsection{Diffusion Inference Visualization} To visually and clearly illustrate the inference process of diffusion, the distributions of prediction errors for the $X$ and $Y$ coordinates of the top-left and bottom-right corners of the bounding boxes at specific reverse steps are shown and visualized separately.

As shown in Figure \ref{Diff_ProcessV}, the reverse step ranges from 0 to 100. It can be observed that as $k$ decreases, the distribution of points converges towards the diagonal line, indicating a gradual reduction in prediction error throughout the denoising process. Additionally, it is noticeable that some points converge to the ground truth more quickly, while others progress more slowly. The former are inferred directly from $X^{obs}$, while the latter are estimated by the neural network. This suggests that the proposed masking inference mechanism contributes to faster convergence while helping to reduce error accumulation.

\subsubsection{Intention Prediction Visualization} 
\begin{figure}[!ht]
\centering
\includegraphics[scale = 0.60]{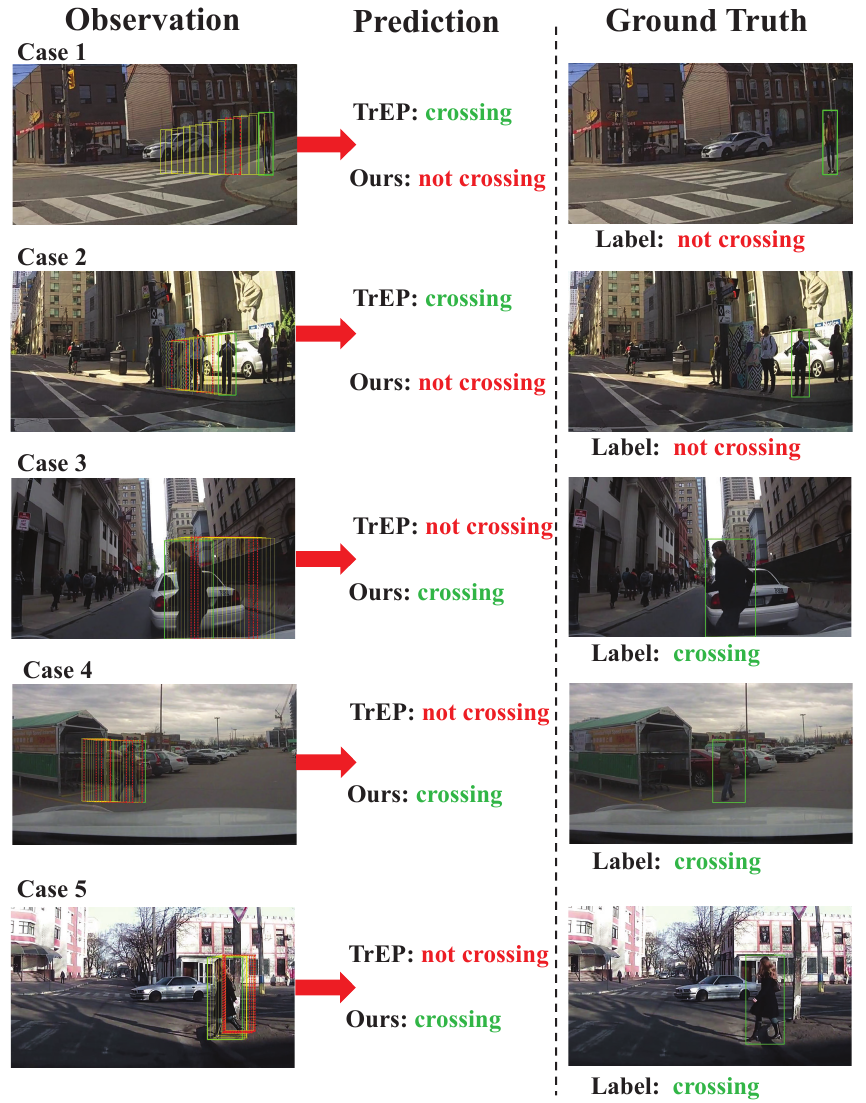}
\caption{
Visualization of intention prediction results under occlusion scenarios. The occlusions are synthetically generated: $(a)$--$(e)$ correspond to occlusion lengths 1 through 5. Cases 1, 2, and 3 represent EO scenarios, while cases 4 and 5 represent PO scenarios. Solid green boxes denote the final positions, dashed red lines represent the missed positions, and yellow boxes indicate the historical positions. }
\centering
\label{Visual1}   
\vspace{-1em}
\end{figure}

To visually assess the proposed model, Figure \ref{Visual1} presents several scenes that are synthetically generated based on the defined masking strategies (EO and PO). In these visualizations, solid green boxes denote the final positions, dashed red boxes indicate masked positions, and yellow boxes represent the observed historical positions. Cases 1, 2, and 4 correspond to EO scenarios, while cases 3 and 5 represent PO scenarios. All cases are compared against the TrEP baseline model.

More specifically, in the relatively simple Cases 1 and 2, where pedestrians remain stationary at the roadside as the vehicle approaches, the proposed method correctly predicts that the pedestrian will not cross, while TrEP mistakenly predicts otherwise. Cases 3 and 4 involve pedestrians crossing with considerably higher uncertainty. In Case 3, the sudden crossing behavior poses a clear potential risk to overall driving safety. In Case 4, the absence of sufficient traffic guidance, such as lanes, zebra crossings, and traffic lights, makes pedestrian movements significantly more difficult to capture. The proposed method accurately aligns with the ground truth labels, while TrEP provides opposite predictions. Case 5 represents a typical urban crossing scenario where a pedestrian moves steadily across the road while multiple vehicles are simultaneously in motion. The predicted results in the figure further underscore our model’s capability to effectively capture occlusion features and foresee potential future movements.

\begin{figure}[!ht]\centering
	\includegraphics[width=8.5cm]{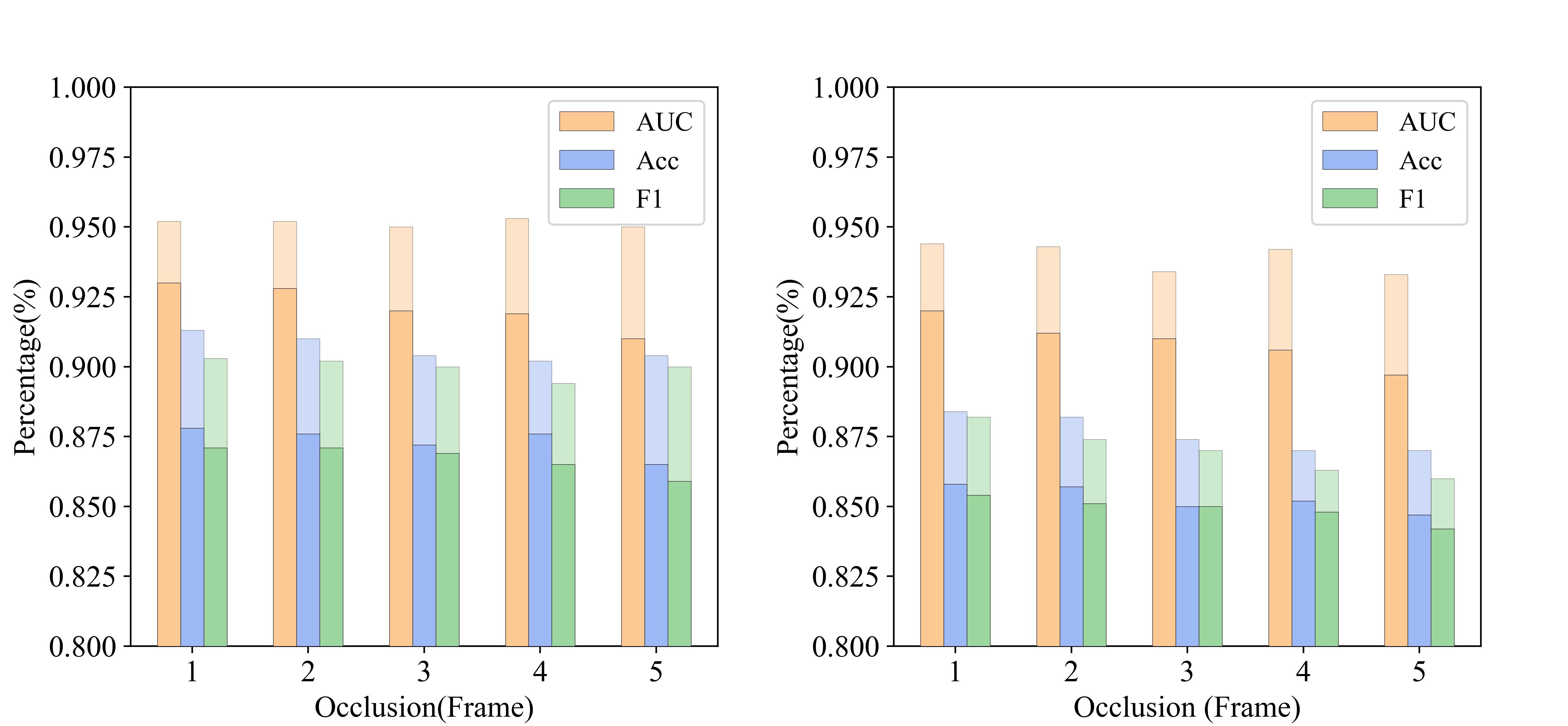}
        \includegraphics[width=8.5cm]{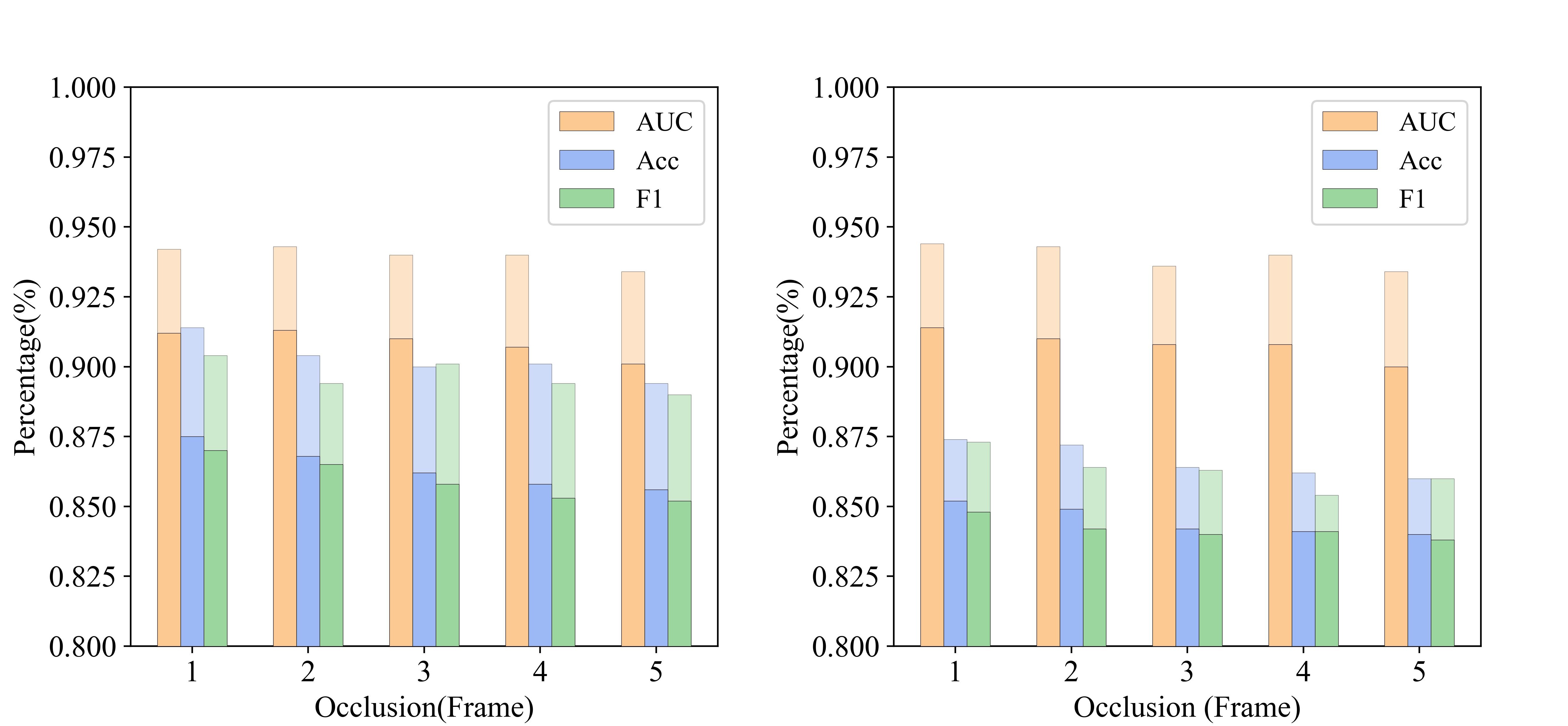}
	\caption{The comparison results of the diffusion mask for EO (top) and PO (bottom) scenarios are shown. On the left, the results for the PIE dataset are presented, while the JAAD dataset results are displayed on the right. Light-colored areas represent the proposed method, whereas dark-colored areas correspond to the compared configuration. }
    \label{Ablation_DMask}
    \vspace{-1em}
\end{figure}
\begin{figure}[!ht]\centering
	\includegraphics[width=8.5cm]{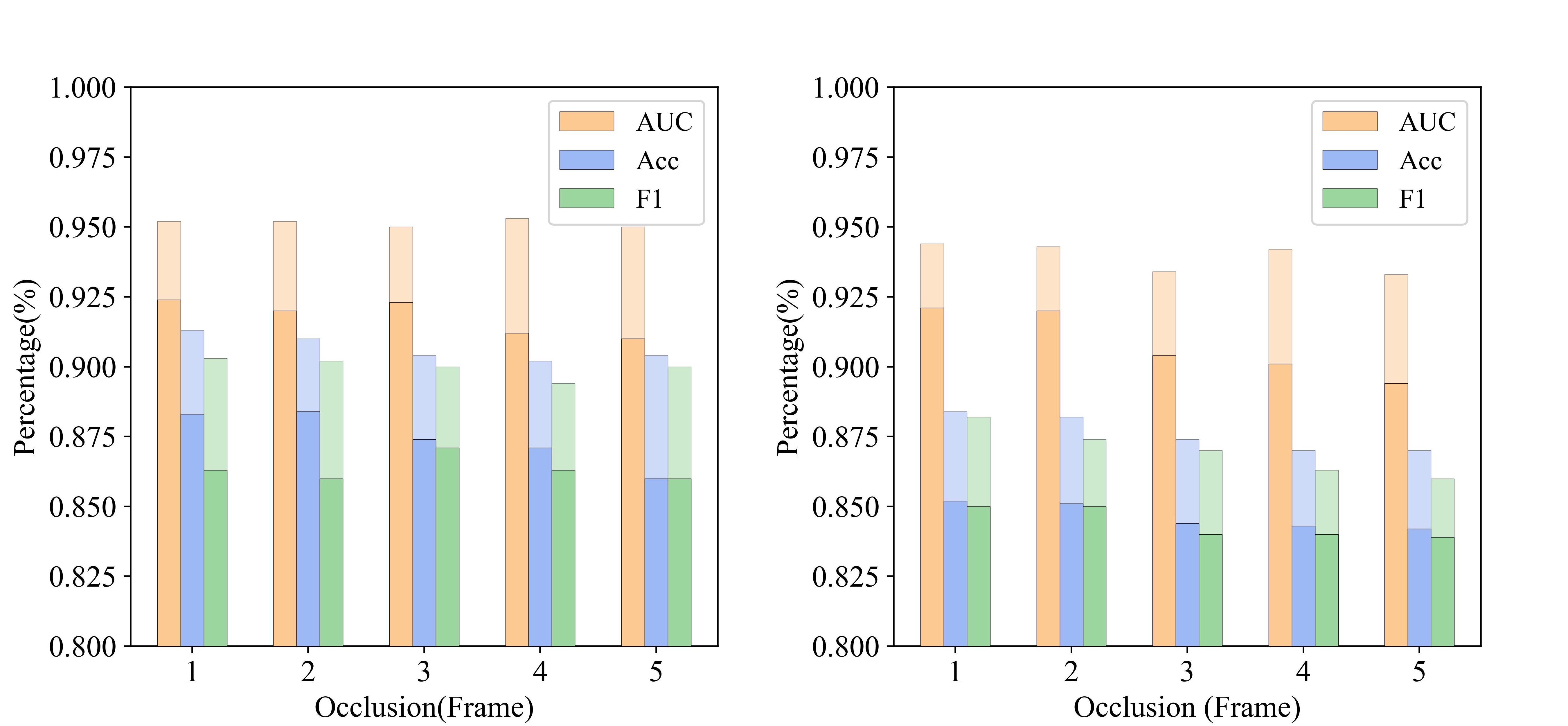}
        \includegraphics[width=8.5cm]{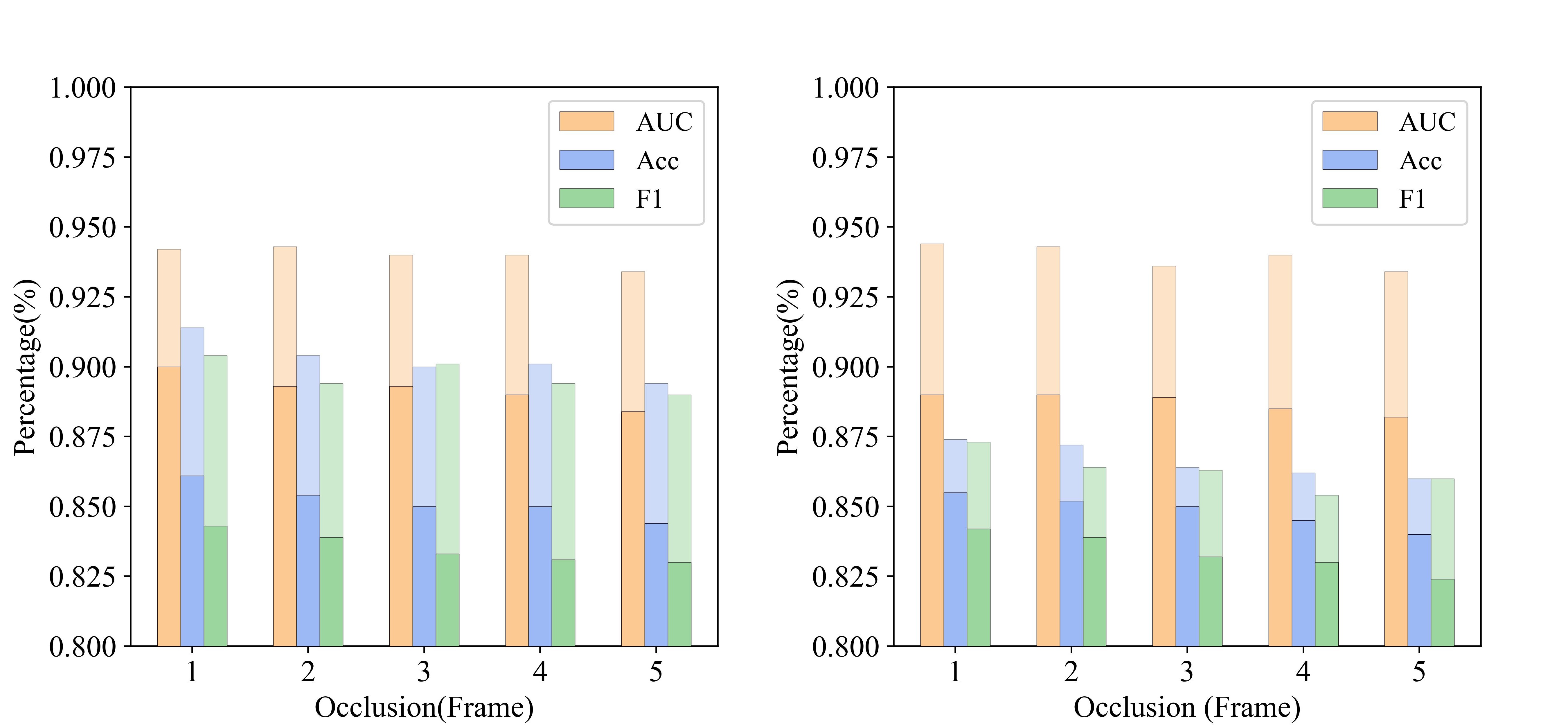}
	\caption{The comparison results of the transformer mask for EO (top) and PO (bottom) scenarios are shown. On the left, the results for the PIE dataset are presented, while the JAAD dataset results are displayed on the right. Light-colored areas represent the proposed method, whereas dark-colored areas correspond to the compared configuration.}
    \label{Ablation_TMask}
\vspace{-1em}
\end{figure}
\subsection{Ablation Study}
In this section, ablation studies are conducted on both the PIE and JAAD datasets to analyze the predictive characteristics of each component of our model in occlusion scenarios. The following aspects are examined.

\subsubsection{Diffusion Mask} In this study, an occlusion mask-guided reference mechanism is proposed during the reverse stage of the diffusion model to enhance prediction performance. To evaluate its impact, an experiment is conducted in which the mask is removed, meaning all features in the subsequent step of the denoising process rely solely on the neural network's estimation rather than being derived from observations $X^{obs}$.

The result is illustrated in Figure \ref{Ablation_DMask}. Areas with deeper colors represent the method without the diffusion mask, while areas with lighter colors correspond to the proposed method using the diffusion mask. Clearly, the model with the diffusion mask consistently outperforms the one without it, leading to significant improvements across all metrics in all occlusion scenarios. For instance, in the EO case with the PIE dataset, maximum increases of 4\%, 4\%, and 3\% are observed for Acc, AUC, and F1, respectively. The results show that incorporating sufficient occlusion patterns during the inference process is crucial for accurately predicting crossing behavior. One possible reason is that deriving results directly from observations, rather than estimating them through the neural network, helps prevent the accumulation of noise prediction errors and facilitates the reverse process, as illustrated in the visualization figure.    

\subsubsection{Transformer Mask } This work proposes an occlusion masking block within the diffusion transformer, aiming to enhance the model's ability to handle occlusion scenarios during the denoising process. A neural network is employed to predict noise features corresponding to the effects of occlusion. In this experiment, the occlusion masking block for prediction is removed, meaning that the features used in the denoising process no longer incorporate occlusion patterns and rely solely on the features of the current state. 

Figure \ref{Ablation_TMask} presents the prediction results for the transformer mask test. Areas with lighter colors represent the proposed method, while areas with darker colors correspond to the cases without the transformer mask. It is evident that the prediction performance of the proposed method surpasses that of the others. Specifically, in PO cases with the PIE, the improvement reaches a maximum of 5\%, 5\%, and 7\% for Acc, AUC, and F1, respectively. These results validate the contribution of the transformer mask in enhancing occlusion-aware semantic understanding for the prediction task.

\subsubsection{Encoder and Decoder Block } In this work, an Occlusion-Aware Spatial-Temporal aggregation method with an AdaLN architecture is employed to integrate observation features with occlusion into the model during the denoising stage. To investigate its impact on prediction performance, we compare the following variations of the block architecture.

Context Conditioning: Instead of learning scale and shift parameters solely from the observation features context, the context features are directly concatenated with the input sequence embedding before being passed through the Occlusion-Aware Spatial-Temporal blocks. This experiment aims to more thoroughly evaluate the impact of this conditional merging method on the model's performance.

Basic Attention Block: To rigorously evaluate the impact of the proposed Occlusion-Aware Spatial-Temporal aggregation mechanism, an experiment is conducted using the AdaLN architecture, where the proposed mechanism is replaced with a standard transformer attention block.
\begin{figure}[!ht]\centering
	\includegraphics[width=8.8cm]{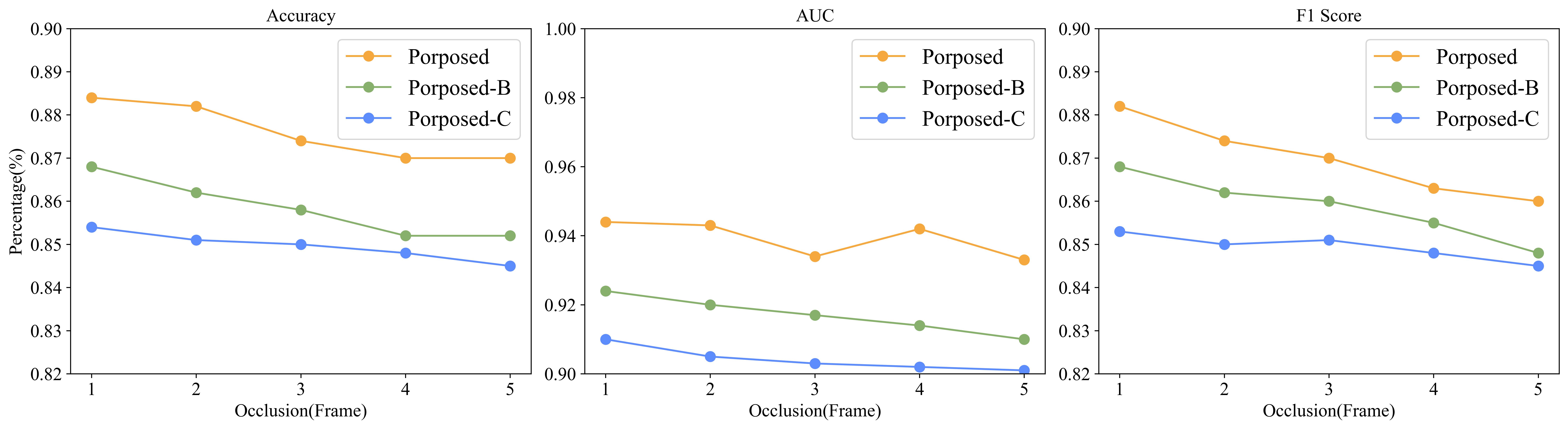}
        \includegraphics[width=8.8cm]{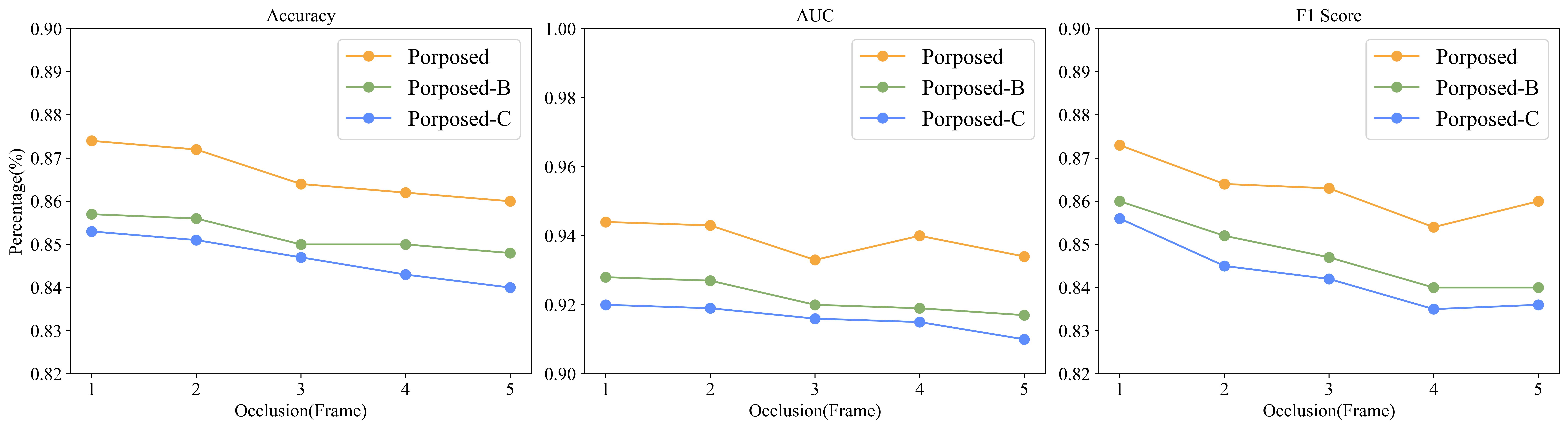}
	\caption{The comparison results of the attention configurations for EO (top) and PO (bottom) scenarios in the JAAD dataset are presented. Proposed-B refers to the proposed method with the basic self-attention block, while Proposed-C denotes the implementation with context conditioning.}
    \label{Attn}
\end{figure}

The experimental results on the JAAD dataset are illustrated in Figure \ref{Attn}. The orange, green, and blue lines represent the proposed method, the basic attention block, and the context conditioning configuration, respectively. It can be observed that the proposed method consistently achieves the highest performance. For example, in the EO case, accuracy, AUC, and F1 increase by a maximum of 3\%, 2\%, and 3\%, respectively, underscoring the importance of the proposed deformable aggregation strategy in achieving optimal results.

\subsubsection{Steps of Diffusion Model} The proposed model uses a transformer-based diffusion process to estimate occluded features, where the number of steps controls denoising. To evaluate its impact, we vary the diffusion steps and report results for EO5 and PO5 on the JAAD dataset.

\begin{figure}[!ht]\centering
	\includegraphics[width=8.5cm]{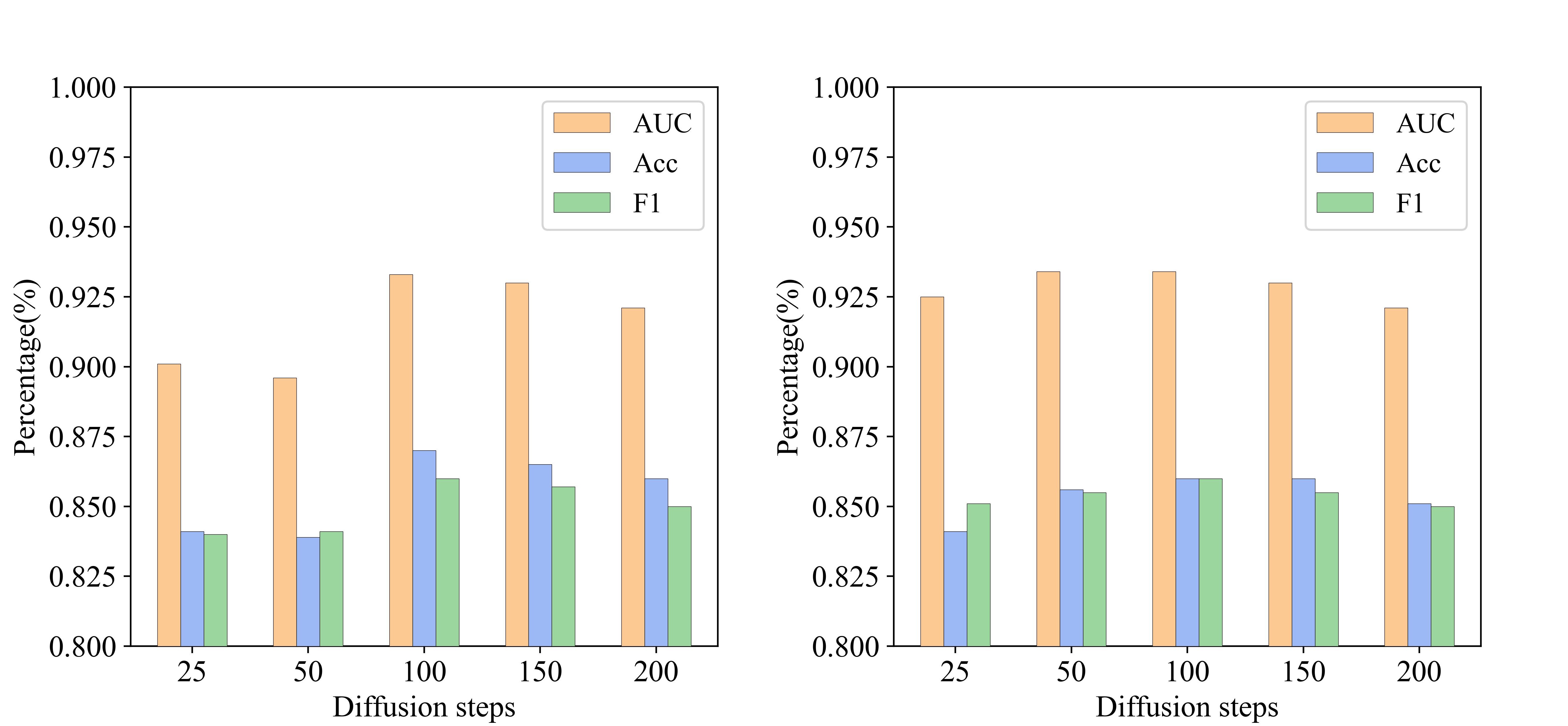}
	\caption{The comparison results of the diffusion steps $K$ for EO5 (left) and PO5 (right) scenarios in the JAAD dataset are presented.}
    \label{Diff_step}
    \vspace{-1em}
\end{figure}

In this test, the diffusion step $K$ is increased from 25 to 200, and the experimental results are shown in Figure \ref{Diff_step}. It can be observed that when $K$ reaches 100 steps, all metrics peak. With smaller diffusion steps, the model may fail to capture the detailed transitions of noise between each step, while with larger diffusion steps, the subtle variations between steps become harder to discriminate due to the limited capacity of the neural network.

\subsubsection{Effectiveness of Feature Recovering } In this work, we propose a masked diffusion model to reconstruct missing motion features caused by occlusions for intention prediction. To evaluate its effectiveness, we conduct an ablation study by removing the diffusion model, meaning that occluded observations are directly utilized for intention prediction by the transformer block without reconstructing complete observation features. The experiment is conducted in the EO5 scenario on PIE, and the results are presented in Table \ref{table_diffu}. The findings indicate that without reconstructing the missing motion features, intention prediction performance declines significantly. This outcome is expected, as the absence of observation features impairs the learning of motion patterns. These results validate the necessity of the feature reconstruction process and demonstrate the effectiveness of the diffusion block.
\vspace{-1em}
\begin{table}[ht]
\centering
\renewcommand{\arraystretch}{0.9}
\setlength{\tabcolsep}{13pt}
\caption{\textsc{Ablation Study of Diffusion Module}} 
\vspace{-1em}
\label{table_diffu}
\begin{tabular}{l c c c} 
\toprule
Methods & Acc $\uparrow$ & AUC $\uparrow$ & F1 $\uparrow$ \\  
\midrule
Proposed w/o Diffusion  & 0.83 & 0.82 & 0.85 \\ 
Proposed  & 0.90 & 0.95 & 0.90 \\ 
\bottomrule
\end{tabular}
\end{table}
\vspace{-1em}


\begin{table}[ht]
\centering
\renewcommand{\arraystretch}{0.9}
\setlength{\tabcolsep}{8pt}
\caption{Ablation Study on Multimodal Inputs. Center (C), Bounding box (B), and Velocity (V).} \label{table_input}
\vspace{-1em}
\begin{tabular}{c c c c c c}  
\toprule
Center & Bounding Box & Velocity & Acc $\uparrow$ & AUC $\uparrow$ & F1 $\uparrow$ \\  
\midrule
            &             & \checkmark & 0.79 & 0.79 & 0.76 \\ 
\checkmark  & \checkmark  &            & 0.86 & 0.86 & 0.84 \\ 
            & \checkmark  & \checkmark & 0.88 & 0.87 & 0.85 \\  
\checkmark  &             & \checkmark & 0.85 & 0.84 & 0.82 \\ 
\checkmark  & \checkmark  & \checkmark & 0.90 & 0.95 & 0.90 \\ 
\bottomrule
\end{tabular}
\vspace{-1em}
\end{table}

\subsubsection{Impact of Different Inputs } In this work, we utilize two input modalities: bounding boxes and ego-vehicle velocity. To further examine potential redundancy, we additionally separate the bounding box center as an independent input. The ablation study is conducted in the EO5 scenario on PIE, with results presented in Table \ref{table_input}. Note that when only velocity input is available, the diffusion model is omitted, as there is no need for observation recovery. In this case, velocity features are directly processed by the intention prediction block.

The results reveal several insights. First, relying solely on ego-vehicle velocity yields the lowest accuracy, underscoring its weak correlation with pedestrian crossing behavior. Furthermore, combining bounding boxes and centers without velocity still produces competitive performance, indicating that bounding boxes capture rich motion cues. When incorporating all three modalities, the highest performance is achieved, highlighting the benefit of multimodal fusion and justifying the inclusion of center information alongside bounding boxes and velocity. Moreover, removing either the center or the bounding box degrades performance, demonstrating that the two provide complementary rather than redundant information.
\begin{figure}[!ht]\centering
	\includegraphics[width=8.8cm]{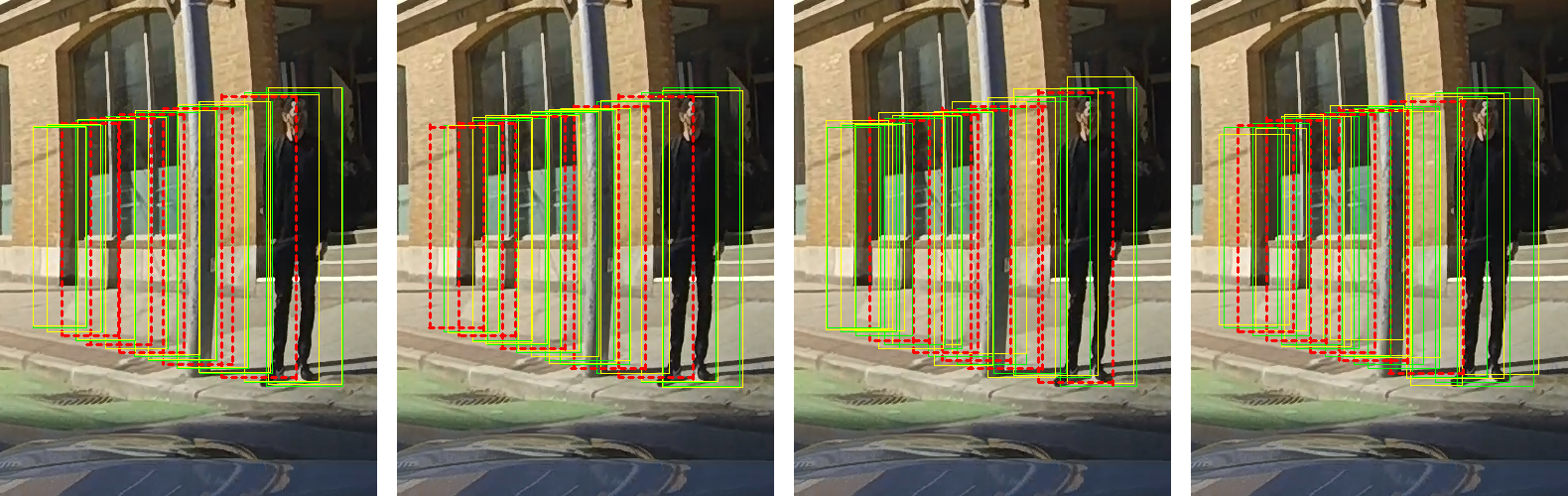}
	\caption{The ground truth bounding box is shown in green, the noised bounding box in yellow, and the occluded bounding box in red. The standard deviations from left to right are 1, 2.5, 5, and 10, respectively. }
    \label{diff_noise}
\vspace{-1em}
\end{figure}
\begin{table}[ht]
\centering
\renewcommand{\arraystretch}{0.9}
\setlength{\tabcolsep}{19pt}
\caption{Perfomance with Noise Observation } 
\vspace{-1em}
\label{table_noise}
\begin{tabular}{c c c c} 
\toprule
 \makecell[c]{Noise Std} & Acc $\uparrow$ & AUC $\uparrow$ & F1 $\uparrow$ \\  
\midrule
0     & 0.90 & 0.95 & 0.90 \\ 
1   & 0.87 & 0.88 & 0.86 \\ 
2.5   & 0.83 & 0.81 & 0.81 \\ 
5   & 0.79 & 0.78 & 0.77 \\
10   & 0.71 & 0.70 & 0.71 \\
\bottomrule
\end{tabular}
\vspace{-1em}
\end{table}

\subsubsection{Effectivenetss of Gate Fusion }
\begin{table}[ht]
\centering
\renewcommand{\arraystretch}{0.9}
\setlength{\tabcolsep}{19pt}
\caption{Ablation Study of Gate Fusion } 
\vspace{-1em}
\label{table_gate}
\begin{tabular}{c c c c} 
\toprule
 \makecell[c]{Noise Std} & Acc $\uparrow$ & AUC $\uparrow$ & F1 $\uparrow$ \\  
\midrule
Gate     & 0.90 & 0.95 & 0.90 \\ 
Concat   & 0.88 & 0.90 & 0.88 \\ 
Average   & 0.85 & 0.87 & 0.87 \\ 
\bottomrule
\end{tabular}
\vspace{-1em}
\end{table}

To evaluate the effectiveness of the proposed gating mechanism, we conducted an ablation study comparing three fusion strategies: concatenation, averaging, and gating in the EO5 on PIE. The results are reported in Table \ref{table_gate}. Compared with concatenation and averaging, the gating mechanism achieves higher accuracy and lower errors, especially under occlusion. This demonstrates that adaptively emphasizing informative modalities while suppressing less reliable ones is beneficial for robust performance. 

\subsubsection{Impact of Noise Observation } In practical applications, in addition to incomplete observations caused by occlusion, signal noise presents another challenge for intelligent systems. To assess prediction performance under such conditions, we introduce noise into the observation sequences with varying standard deviations from a Gaussian distribution as shown in Figure \ref{diff_noise}. The experiment is conducted in the EO5 scenario on PIE, and the results are summarized in Table \ref{table_noise}. As expected, the findings indicate that higher noise levels degrade prediction performance, as greater deviations disrupt motion comprehension for intention prediction. From this perspective, enhancing robustness against noise contamination remains an important direction for further exploration.

\begin{figure}[ht]
    \centering
    \subfloat{\includegraphics[width=0.24\textwidth]{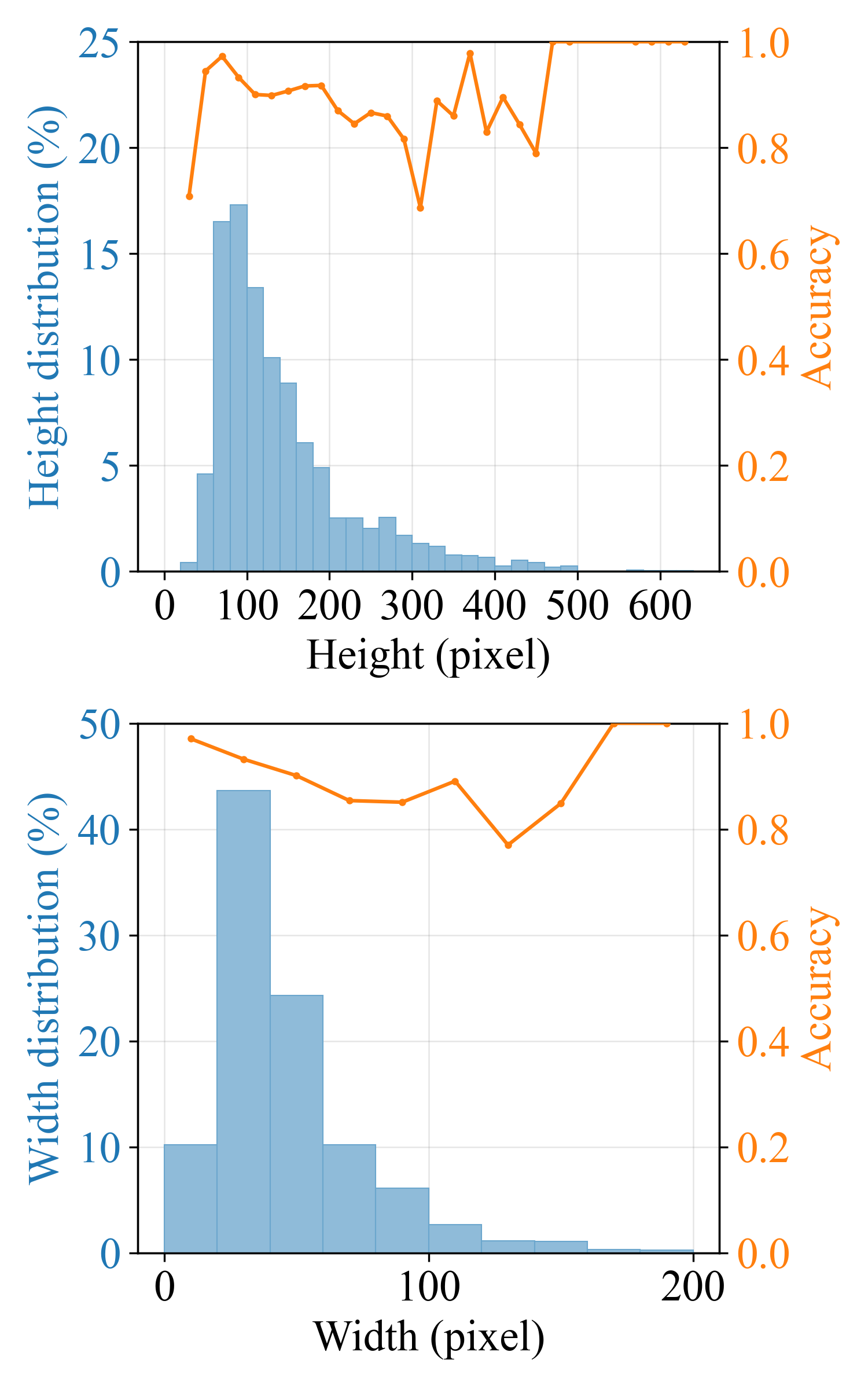}} 
    \hspace{-0.012\textwidth}
    \subfloat{\includegraphics[width=0.24\textwidth]{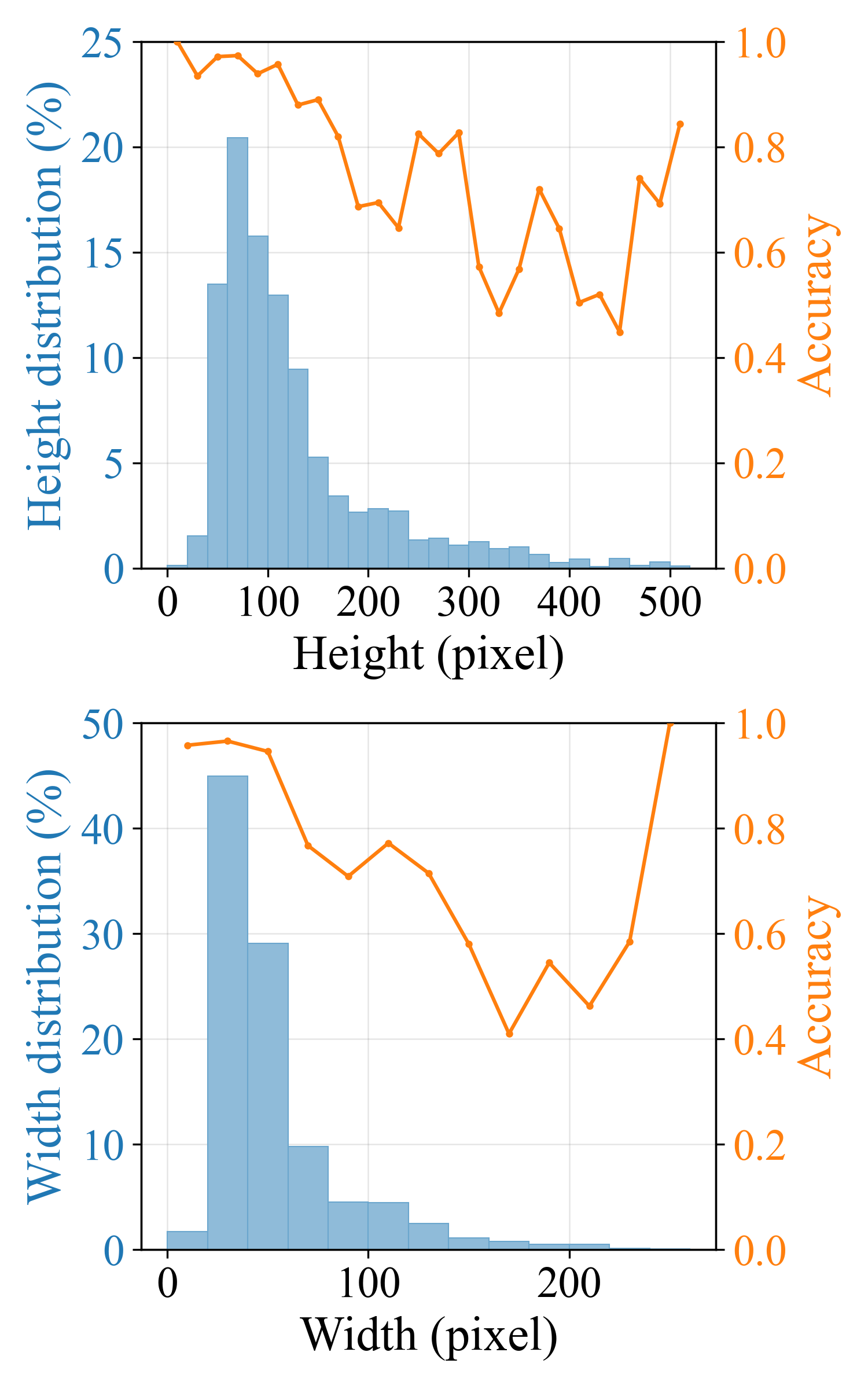}} 
    \caption{Bounding box size distributions (blue bars, left axis) and corresponding accuracy trends (lines, right axis) are shown. The top and bottom rows represent bounding box height and width, respectively, with PIE on the left and JAAD on the right. }
    \label{Dist}
    \vspace{-1em}
\end{figure}
\subsubsection{Impact of Scale} The impact of bounding box size on prediction performance is illustrated in Figure~\ref{Dist}. In PIE, most bounding box heights fall within the range of $[60, 120]$, and widths within $[20, 60]$, which is similar to the distribution observed in JAAD. High accuracy performance is also observed in regions with large bounding box sizes. For instance, in PIE, high accuracy is achieved when the height exceeds 500 and the width exceeds 160. This may be because larger bounding boxes often correspond to pedestrians at closer distances, where motion patterns are more explicit and thus more informative for inferring crossing intentions than those observed at greater distances with small sizes. 

\subsection{Occlusion Reconstruction Performance}
As the initial stage of the framework, accurately estimating occluded bounding boxes is particularly essential for reliable downstream prediction. We adopt the standard Average Displacement Error (ADE), defined as the mean $L_2$ distance between the predicted bounding boxes and the ground truth bounding boxes across all frames and sequences:
\begin{equation}
ADE= \frac{1}{N \times T} \sum_{i=1}^N \sum_{t=1}^{T} \parallel{\hat{b}_t^i - b_t^i} \parallel ^2.
\end{equation}  

Table \ref{Table 1} reports the reconstruction performance under occlusion lengths from one to five frames and two occlusion patterns EO and PO. Given that the pixel size of the images is $1920 \times 1080$, the results show that, even in scenarios of significant non-observability, the diffusion model effectively preserves its ability to accurately reconstruct occluded traffic information, maintaining the error around 15 pixels.

\begin{table}[ht]
\centering
\setlength{\tabcolsep}{12 pt} 
\renewcommand{\arraystretch}{0.9} 
\caption{Recovery Performance under Occlusions}
\vspace{-1em}
\label{Table 1}
\begin{tabular}{@{}ccccc@{}}
\toprule
\multirow{2}{*}{\begin{tabular}[c]{@{}c@{}}Occlusion\\ Length\end{tabular}} & \multicolumn{2}{c}{PIE} & \multicolumn{2}{c}{JAAD} \\ \cmidrule(lr){2-3} \cmidrule(lr){4-5}
          & Bbox $\downarrow $          & Center $\downarrow $       & Bbox $\downarrow $       & Center $\downarrow $       \\     \midrule 
EO, 1     & 15.752        & 14.709       & 14.487      & 14.017       \\ 
EO, 2     & 15.672        & 14.685       & 14.231      & 13.759        \\
EO, 3     & 15.785        & 14.801       & 14.220      & 13.764        \\ 
EO, 4     & 15.864        & 14.737       & 14.367      & 13.886        \\
EO, 5     & 15.736        & 15.736       & 14.331      & 13.873        \\     \midrule 
PO, 1     & 15.720        & 14.635       & 14.560      & 14.052        \\ 
PO, 2     & 15.632        & 14.567       & 14.739      & 14.247        \\
PO, 3     & 15.514        & 14.641       & 14.448      & 13.964        \\ 
PO, 4     & 15.646        & 14.638       & 14.288      & 13.794         \\
PO, 5     & 15.663        & 14.793       & 14.557      & 14.093         \\ 
\bottomrule
\end{tabular}
\end{table}
\vspace{-1em}

\subsection{Failure Cases}
\begin{figure}[ht]
    \centering
    \subfloat{\includegraphics[width=0.48\textwidth]{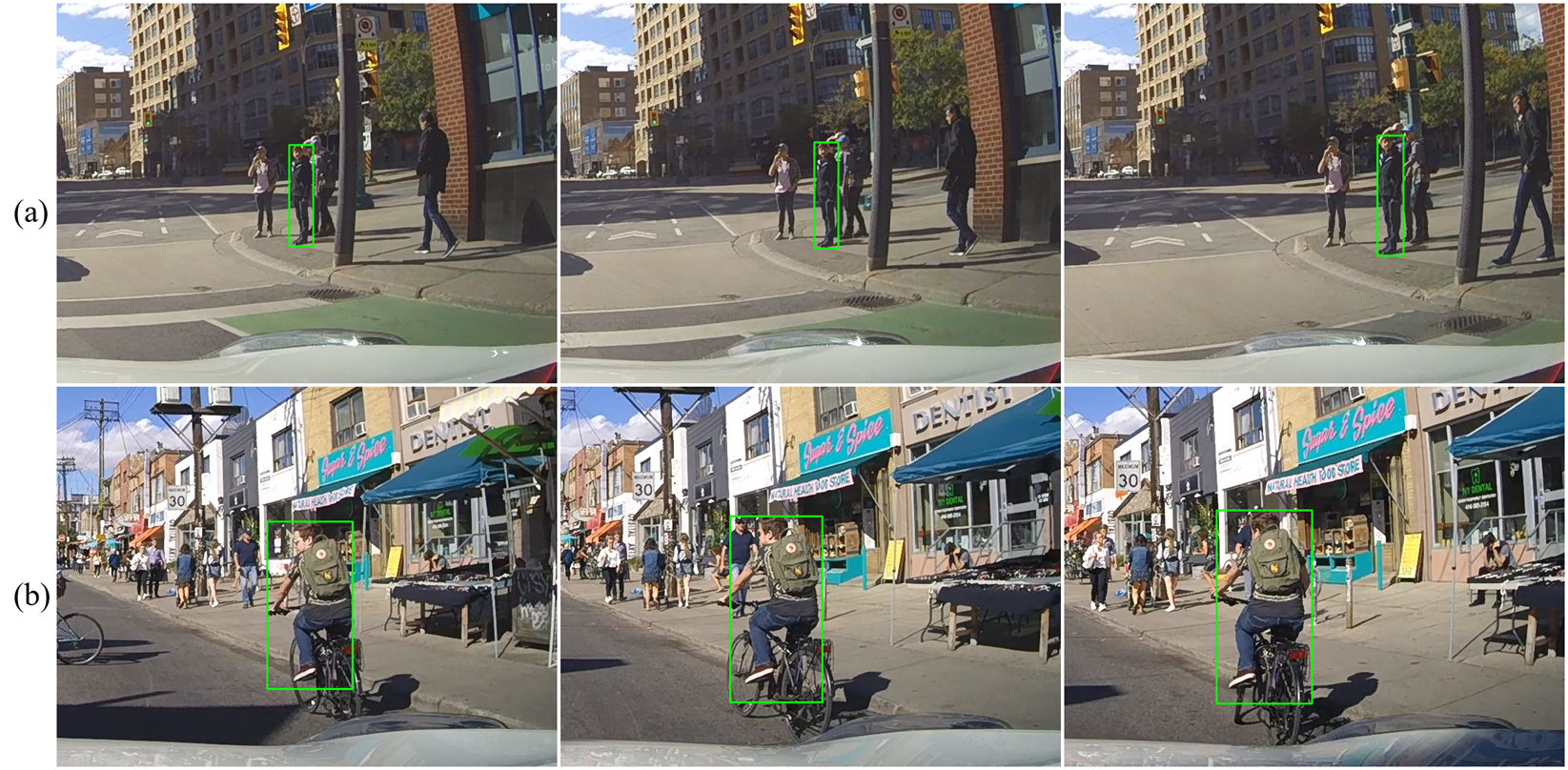}} 
    \caption{   
    Failure case study. Two representative failure cases are illustrated in (a) and (b). In both instances, although the pedestrians show no intention to cross, the model incorrectly predicted their intentions.
    }
    \label{Fail}
    \vspace{-1em}
\end{figure}

In Figure \ref{Fail}, two representative failure cases are illustrated. In the corner scene (a), ego-vehicle motion induces background parallax: a stationary pedestrian appears to drift relative to curb lines and buildings. As the vehicle approaches, the bounding box expands and shifts slightly, mimicking walking onset. Without scene-level priors such as curb geometry or ego-motion compensation, the model misinterprets this illusion as true pedestrian motion. In the crossing scene (b), the pedestrian’s motion tendency is abruptly interrupted by a sudden behavioral change. The pedestrian steps forward just as the ego-vehicle approaches, creating a sharp transition that alters the expected trajectory. Without temporal priors on sudden intent shifts or richer scene-level cues such as pedestrian-vehicle interaction dynamics, the model fails to capture this abrupt change and produces an incorrect prediction.

\section{Conclusion}
In this work, we propose ODM, an occlusion-aware diffusion model for pedestrian intention prediction in occlusion scenarios, specifically applicable when an intelligent vehicle's onboard sensors fail to capture pedestrian states in complex environments. The study highlights the importance of reconstructing missing motion features for accurate intention prediction. To address this, we design a diffusion model with an occlusion-guided masking mechanism that promotes reverse process convergence, and an occlusion-masked diffusion transformer with an Occlusion-Aware Spatio-Temporal encoder–decoder to capture correlations between occlusion patterns and pedestrian motion. An occlusion masking block further enhances contextual representations. Extensive experiments and ablation studies on multiple datasets confirm that ODM consistently outperforms state-of-the-art methods under various occlusion levels, demonstrating strong robustness and generalization across challenging visual conditions.

We plan to enhance the model’s adaptability to diverse and unseen environments by leveraging transfer learning and domain adaptation techniques, enabling effective deployment in new environments with minimal retraining. In addition, we will explore lightweight network architectures and knowledge distillation strategies to reduce inference latency and computational overhead, paving the way for real-time pedestrian intention prediction in resource-constrained autonomous systems.

\bibliographystyle{IEEEtran}
\bibliography{refs}

\begin{thebibliography}{10}
\providecommand{\url}[1]{#1}
\csname url@samestyle\endcsname
\providecommand{\newblock}{\relax}
\providecommand{\bibinfo}[2]{#2}
\providecommand{\BIBentrySTDinterwordspacing}{\spaceskip=0pt\relax}
\providecommand{\BIBentryALTinterwordstretchfactor}{4}
\providecommand{\BIBentryALTinterwordspacing}{\spaceskip=\fontdimen2\font plus
\BIBentryALTinterwordstretchfactor\fontdimen3\font minus
  \fontdimen4\font\relax}
\providecommand{\BIBforeignlanguage}[2]{{%
\expandafter\ifx\csname l@#1\endcsname\relax
\typeout{** WARNING: IEEEtran.bst: No hyphenation pattern has been}%
\typeout{** loaded for the language `#1'. Using the pattern for}%
\typeout{** the default language instead.}%
\else
\language=\csname l@#1\endcsname
\fi
#2}}
\providecommand{\BIBdecl}{\relax}
\BIBdecl

\bibitem{ref3}
S.~Eiffert, H.~Kong, N.~Pirmarzdashti, and S.~Sukkarieh, ``Path planning in
  dynamic environments using generative rnns and monte carlo tree search,'' in
  \emph{Proc. IEEE Int. Conf. Robot. Autom. (ICRA)}, May. 2020, pp.
  10\,263--10\,269.

\bibitem{ref94}
Z.~Yin, T.~Lai, S.~Khan, J.~Jacob, Y.~Li, and F.~Ramos, ``Stein movement
  primitives for adaptive multi-modal trajectory generation,'' in \emph{2024
  IEEE/RSJ International Conference on Intelligent Robots and Systems (IROS)},
  2024, pp. 11\,901--11\,908.

\bibitem{ref2}
Y.~Liu, Y.~Zhang, K.~Li, Y.~Qiao, S.~Worrall, Y.-F. Li, and H.~Kong,
  ``Knowledge-aware graph transformer for pedestrian trajectory prediction,''
  in \emph{Proc. IEEE Intell. Transp. Syst. Conf. (ITSC)}, Sep. 2023, pp.
  4360--4366.

\bibitem{ref1}
K.~Li, M.~Shan, K.~Narula, S.~Worrall, and E.~Nebot, ``Socially aware crowd
  navigation with multimodal pedestrian trajectory prediction for autonomous
  vehicles,'' in \emph{Proc. IEEE Intell. Transp. Syst. Conf. (ITSC)}, Sep.
  2020, pp. 1--8.

\bibitem{ref4}
A.~Alahi, K.~Goel, V.~Ramanathan, A.~Robicquet, L.~Fei-Fei, and S.~Savarese,
  ``Social {LSTM}: {H}uman trajectory prediction in crowded spaces,'' in
  \emph{Proc. IEEE Conf. Comput. Vis. Pattern Recognit. (CVPR)}, Jun. 2016, pp.
  961--971.

\bibitem{ref27}
J.~Fang, F.~Wang, J.~Xue, and T.-S. Chua, ``Behavioral intention prediction in
  driving scenes: A survey,'' \emph{IEEE Trans. Intell. Transp. Syst.},
  vol.~25, no.~8, pp. 8334--8355, Aug, 2024.

\bibitem{ref9}
I.~Kotseruba, A.~Rasouli, and J.~K. Tsotsos, ``Do they want to cross?
  understanding pedestrian intention for behavior prediction,'' in \emph{Proc.
  IEEE Intell. Vehicles Symp. (IV)}, Oct. 2020, pp. 1688--1693.

\bibitem{ref10}
J.~Lorenzo, I.~Parra, F.~Wirth, C.~Stiller, D.~F. Llorca, and M.~A. Sotelo,
  ``Rnn-based pedestrian crossing prediction using activity and pose-related
  features,'' in \emph{Proc. IEEE Intell. Vehicles Symp. (IV)}, Oct. 2020, pp.
  1801--1806.

\bibitem{ref11}
A.~Rasouli, M.~Rohani, and J.~Luo, ``Bifold and semantic reasoning for
  pedestrian behavior prediction,'' in \emph{Proc. IEEE Int. Conf. Comput. Vis.
  (ICCV)}, Oct. 2021, pp. 15\,600--15\,610.

\bibitem{ref73}
L.~Achaji, J.~Moreau, T.~Fouqueray, F.~Aioun, and F.~Charpillet, ``Is attention
  to bounding boxes all you need for pedestrian action prediction?'' in
  \emph{Proc. IEEE Intell. Vehicles Symp. (IV)}, Jun. 2022, pp. 895--902.

\bibitem{ref14}
C.~Hu, R.~Niu, Y.~Lin, B.~Yang, H.~Chen, B.~Zhao, and X.~Zhang, ``Probabilistic
  trajectory prediction of vulnerable road user using multimodal inputs,''
  \emph{IEEE Trans. Intell. Transp. Syst.}, vol.~26, no.~2, pp. 2679--2689,
  Feb. 2025.

\bibitem{ref12}
A.~Singh and U.~Suddamalla, ``Multi-input fusion for practical pedestrian
  intention prediction,'' in \emph{Proc. IEEE Int. Conf. Comput. Vis. (ICCV)},
  Oct. 2021, pp. 2304--2311.

\bibitem{ref7}
F.~Giuliari, I.~Hasan, M.~Cristani, and F.~Galasso, ``Transformer networks for
  trajectory forecasting,'' in \emph{Proc. Int. Conf. Pattern Recognit.
  (ICPR)}, Jan. 2021, pp. 10\,335--10\,342.

\bibitem{ref8}
K.~Li, M.~Shan, S.~Worrall, and E.~Nebot, ``Crowd prediction and autonomous
  navigation with partial observations,'' in \emph{Proc. IEEE Intell. Transp.
  Syst. Conf. (ITSC)}, Oct, 2022, pp. 1829--1836.

\bibitem{ref90}
I.~Kotseruba, A.~Rasouli, and J.~K. Tsotsos, ``Benchmark for evaluating
  pedestrian action prediction,'' in \emph{IEEE Winter Conf. Appl. Comput. Vis.
  (WACV)}, Jun. 2021, pp. 1257--1267.

\bibitem{ref74}
Z.~Zhang, R.~Tian, and Z.~Ding, ``Trep: Transformer-based evidential prediction
  for pedestrian intention with uncertainty,'' in \emph{Proc. AAAI Conf. Artif.
  Intell.}, vol.~37, no.~3, Jun. 2023, pp. 3534--3542.

\bibitem{ref60}
W.~Choi and S.~Savarese, ``Understanding collective activitiesof people from
  videos,'' \emph{IEEE Trans. Pattern Anal. Mach. Intell.}, vol.~36, no.~6, pp.
  1242--1257, Jun. 2014.

\bibitem{ref61}
V.~Karasev, A.~Ayvaci, B.~Heisele, and S.~Soatto, ``Intent-aware long-term
  prediction of pedestrian motion,'' in \emph{Proc. IEEE Int. Conf. Robot.
  Autom. (ICRA)}, May. 2016, pp. 2543--2549.

\bibitem{ref62}
J.~F.~P. Kooij, N.~Schneider, F.~Flohr, and D.~M. Gavrila, ``Context-based
  pedestrian path prediction,'' in \emph{Proc. Eur. Conf. Comput. Vis. (ECCV)},
  Sep. 2014, pp. 618--633.

\bibitem{ref63}
A.~Rasouli, I.~Kotseruba, T.~Kunic, and J.~Tsotsos, ``Pie: A large-scale
  dataset and models for pedestrian intention estimation and trajectory
  prediction,'' in \emph{Proc. IEEE Int. Conf. Comput. Vis. (ICCV)}, Oct. 2019,
  pp. 6261--6270.

\bibitem{ref64}
A.~Rasouli, I.~Kotseruba, and J.~K. Tsotsos, ``Are they going to cross? a
  benchmark dataset and baseline for pedestrian crosswalk behavior,'' in
  \emph{Proc. IEEE Int. Conf. Comput. Vis. Workshops (ICCVW)}, Oct. 2017, pp.
  206--213.

\bibitem{ref92}
A.~Bhattacharyya, M.~Fritz, and B.~Schiele, ``Long-term on-board prediction of
  people in traffic scenes under uncertainty,'' in \emph{Proc. IEEE Conf.
  Comput. Vis. Pattern Recognit. (CVPR)}, Jun. 2018, pp. 4194--4202.

\bibitem{ref91}
A.~Rasouli, I.~Kotseruba, and J.~K. Tsotsos, ``Pedestrian action anticipation
  using contextual feature fusion in stacked rnns,'' in \emph{Proc. Brit. Mach.
  Vis. Conf. (BMVC)}, Sep. 2019.

\bibitem{ref65}
Z.~Fang and A.~M. López, ``Is the pedestrian going to cross? answering by 2d
  pose estimation,'' in \emph{Proc. IEEE Intell. Vehicles Symp. (IV)}, Jun.
  2018, pp. 1271--1276.

\bibitem{ref66}
S.~Zhang, M.~Abdel-Aty, Y.~Wu, and O.~Zheng, ``Pedestrian crossing intention
  prediction at red-light using pose estimation,'' \emph{IEEE Trans. Intell.
  Transp. Syst.}, vol.~23, no.~3, pp. 2331--2339, Mar. 2022.

\bibitem{ref68}
B.~Liu, E.~Adeli, Z.~Cao, K.-H. Lee, A.~Shenoi, A.~Gaidon, and J.~C. Niebles,
  ``Spatiotemporal relationship reasoning for pedestrian intent prediction,''
  \emph{IEEE Robot. Autom. Lett.}, vol.~5, no.~2, pp. 3485--3492, Feb. 2020.

\bibitem{ref69}
X.~Zhang, P.~Angeloudis, and Y.~Demiris, ``St crossingpose: A spatial-temporal
  graph convolutional network for skeleton-based pedestrian crossing intention
  prediction,'' \emph{IEEE Trans. Intell. Transp. Syst.}, vol.~23, no.~11, pp.
  20\,773--20\,782, Nov. 2022.

\bibitem{ref70}
P.~R.~G. Cadena, Y.~Qian, C.~Wang, and M.~Yang, ``Pedestrian graph +: A fast
  pedestrian crossing prediction model based on graph convolutional networks,''
  \emph{IEEE Trans. Intell. Transp. Syst.}, vol.~23, no.~11, pp.
  21\,050--21\,061, Nov. 2022.

\bibitem{ref71}
A.~Rasouli, T.~Yau, M.~Rohani, and J.~Luo, ``Multi-modal hybrid architecture
  for pedestrian action prediction,'' in \emph{Proc. IEEE Intell. Vehicles
  Symp. (IV)}, Jun. 2022, pp. 91--97.

\bibitem{ref72}
D.~Yang, H.~Zhang, E.~Yurtsever, K.~A. Redmill, and {\"U}.~{\"O}zg{\"u}ner,
  ``Predicting pedestrian crossing intention with feature fusion and
  spatio-temporal attention,'' \emph{IEEE Trans. Intell. Vehicles}, vol.~7,
  no.~2, pp. 221--230, Jun. 2022.

\bibitem{ref6}
A.~Vaswani, N.~Shazeer, N.~Parmar, J.~Uszkoreit, L.~Jones, A.~N. Gomez,
  {\L}.~Kaiser, and I.~Polosukhin, ``Attention is all you need,'' in
  \emph{Proc. Conf. Neural Inf. Process. Syst. (NIPS)}, vol.~30, 2017, pp.
  1--11.

\bibitem{ref75}
X.~Chen, S.~Zhang, J.~Li, and J.~Yang, ``Pedestrian crossing intention
  prediction based on cross-modal transformer and uncertainty-aware multi-task
  learning for autonomous driving,'' \emph{IEEE Trans. Intell. Transp. Syst.},
  vol.~25, no.~9, pp. 12\,538--12\,549, Sep. 2024.

\bibitem{ref80}
J.~Sohl-Dickstein, E.~Weiss, N.~Maheswaranathan, and S.~Ganguli, ``Deep
  unsupervised learning using nonequilibrium thermodynamics,'' in \emph{Proc.
  Int. Conf. Mach. Learn. (ICML)}, Jun. 2015, pp. 2256--2265.

\bibitem{ref81}
J.~Ho, A.~Jain, and P.~Abbeel, ``Denoising diffusion probabilistic models,'' in
  \emph{Proc. Adv. Neural Inf. Process. Syst. (NIPS)}, vol.~33, Dec. 2020, pp.
  6840--6851.

\bibitem{ref82}
P.~Dhariwal and A.~Nichol, ``Diffusion models beat gans on image synthesis,''
  in \emph{Proc. Adv. Neural Inf. Process. Syst. (NIPS)}, vol.~34, Dec. 2021,
  pp. 8780--8794.

\bibitem{ref95}
Z.~Yin, T.~Lai, L.~Barcelos, J.~Jacob, Y.~Li, and F.~Ramos, ``Diverse motion
  planning with stein diffusion trajectory inference,'' in \emph{2025 IEEE
  International Conference on Robotics and Automation (ICRA)}, 2025, pp.
  15\,610--15\,616.

\bibitem{ref83}
A.~Q. Nichol and P.~Dhariwal, ``Improved denoising diffusion probabilistic
  models,'' in \emph{Proc. Int. Conf. Mach. Learn. (ICML)}, Jun. 2021, pp.
  8162--8171.

\bibitem{ref88}
J.~Ho and T.~Salimans, ``Classifier-free diffusion guidance,'' in \emph{Proc.
  Adv. Neural Inf. Process. Syst. Workshop DGMs Appl.}, Dec. 2021.

\bibitem{ref89}
R.~Rombach, A.~Blattmann, D.~Lorenz, P.~Esser, and B.~Ommer, ``High-resolution
  image synthesis with latent diffusion models,'' in \emph{Proc. IEEE Conf.
  Comput. Vis. Pattern Recognit. (CVPR)}, Jun. 2022, pp. 10\,684--10\,695.

\bibitem{ref84}
F.~Bao, S.~Nie, K.~Xue, Y.~Cao, C.~Li, H.~Su, and J.~Zhu, ``All are worth
  words: A vit backbone for diffusion models,'' in \emph{Proc. IEEE Conf.
  Comput. Vis. Pattern Recognit. (CVPR)}, Jun. 2023, pp. 22\,669--22\,679.

\bibitem{ref85}
W.~Peebles and S.~Xie, ``Scalable diffusion models with transformers,'' in
  \emph{Proc. IEEE Int. Conf. Comput. Vis. (ICCV)}, Oct. 2023, pp. 4195--4205.

\bibitem{ref86}
S.~Gao, P.~Zhou, M.-M. Cheng, and S.~Yan, ``Masked diffusion transformer is a
  strong image synthesizer,'' in \emph{Proc. IEEE Int. Conf. Comput. Vis.
  (ICCV)}, Oct. 2023, pp. 23\,164--23\,173.

\bibitem{ref76}
T.~Gu, G.~Chen, J.~Li, C.~Lin, Y.~Rao, J.~Zhou, and J.~Lu, ``Stochastic
  trajectory prediction via motion indeterminacy diffusion,'' in \emph{Proc.
  IEEE Conf. Comput. Vis. Pattern Recognit. (CVPR)}, Jun. 2022, pp.
  17\,113--17\,122.

\bibitem{ref77}
K.~Chen, X.~Chen, Z.~Yu, M.~Zhu, and H.~Yang, ``Equidiff: A conditional
  equivariant diffusion model for trajectory prediction,'' in \emph{Proc. IEEE
  Intell. Transp. Syst. Conf. (ITSC)}, Sep. 2023, pp. 746--751.

\bibitem{ref78}
S.~Lian, B.~Zhou, S.~Hu, J.~Hu, G.~Wang, J.~Escribano, X.~Na, and S.~Jin,
  ``Enhanced multimodal trajectory prediction for autonomous vehicles using
  advanced diffusion model techniques,'' in \emph{Proc. IEEE Intell. Vehicles
  Symp. (IV)}, Jun. 2024, pp. 484--489.

\bibitem{ref79}
C.~Jiang, A.~Cornman, C.~Park, B.~Sapp, Y.~Zhou, D.~Anguelov \emph{et~al.},
  ``Motiondiffuser: Controllable multi-agent motion prediction using
  diffusion,'' in \emph{Proc. IEEE Conf. Comput. Vis. Pattern Recognit.
  (CVPR)}, Jun. 2023, pp. 9644--9653.

\bibitem{ref87}
Y.~Choi, R.~C. Mercurius, S.~M.~A. Shabestary, and A.~Rasouli, ``Dice: Diverse
  diffusion model with scoring for trajectory prediction,'' in \emph{Proc. IEEE
  Intell. Vehicles Symp. (IV)}, Jun. 2024, pp. 3023--3029.

\bibitem{ref93}
J.~Ma, Z.~Zhao, X.~Yi, J.~Chen, L.~Hong, and E.~H. Chi, ``Modeling task
  relationships in multi-task learning with multi-gate mixture-of-experts,'' in
  \emph{Proc. ACM SIGKDD Int. Conf. Knowl. Discov. Data Mining}, 2018, p.
  1930–1939.

\bibitem{ref58}
Z.~Xia, X.~Pan, S.~Song, L.~E. Li, and G.~Huang, ``Vision transformer with
  deformable attention,'' in \emph{Proc. IEEE Conf. Comput. Vis. Pattern
  Recognit. (CVPR)}, Jun. 2022, pp. 4784--4793.

\bibitem{ref57}
J.~Carreira and A.~Zisserman, ``Quo vadis, action recognition? a new model and
  the kinetics dataset,'' in \emph{Proc. IEEE Conf. Comput. Vis. Pattern
  Recognit. (CVPR)}, Jun. 2017, pp. 4724--4733.

\end{thebibliography}

\vspace{-100mm}
\begin{IEEEbiography}[{\includegraphics[width=1in,height=1.25in,clip,keepaspectratio]{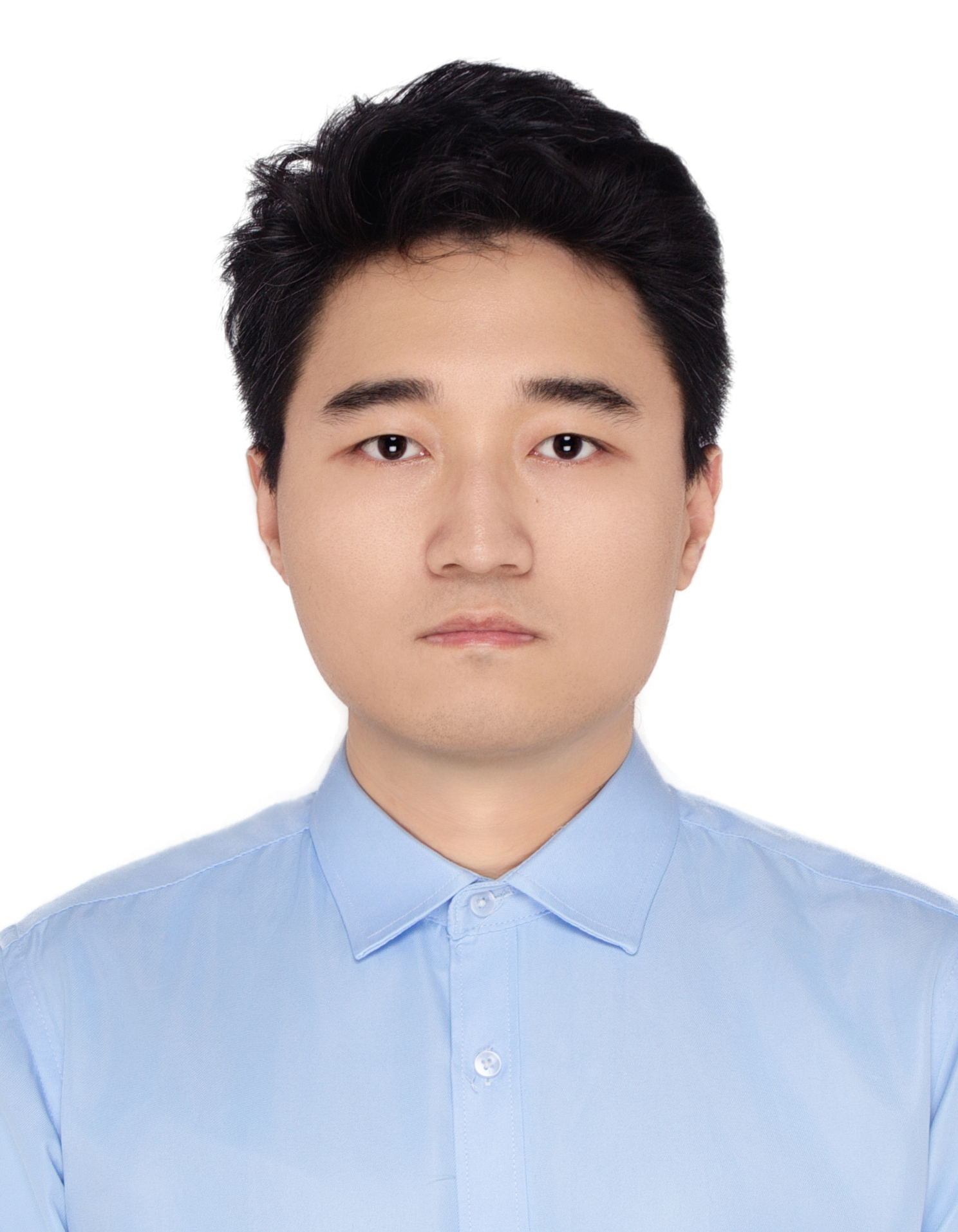}}]{Yu Liu}
received the Bachelor's degree in Process Equipment and Control Engineering from Chongqing University of Technology, Chongqing, China, and the Master's degree in Aerospace Engineering from Nagoya University, Nagoya, Japan. He is currently pursuing the Ph.D. degree with the Department of Mechanical Engineering, City University of Hong Kong(CityU), Hong Kong SAR, China, and also with the Shenzhen Key Laboratory of Control Theory and Intelligent Systems, Southern University of Science and Technology (SUSTech), Shenzhen, China. His current research interests include deep learning, motion prediction, and intelligent vehicles.
\end{IEEEbiography}

\vspace{-100mm}

\begin{IEEEbiography}[{\includegraphics[width=1in,height=1.25in,clip,keepaspectratio]{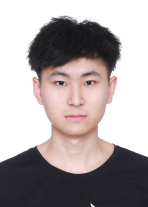}}]{Zhijie Liu}
    received the B.S. degree in Measurement and Control Technology and Instrument from Harbin Engineering University, Harbin, China in 2023. He is currently pursuing the M.S. degree at the School of System Design and Intelligent Manufacturing, Southern University of Science and Technology(SUSTech), Shenzhen, China. His research interests include  nonlinear control, underactuated systems,optimal control, observer design, and high-order fully actuated system approaches.
\end{IEEEbiography}

\vspace{-100mm}
\begin{IEEEbiography}[{\includegraphics[width=1in,height=1.25in,clip,keepaspectratio]{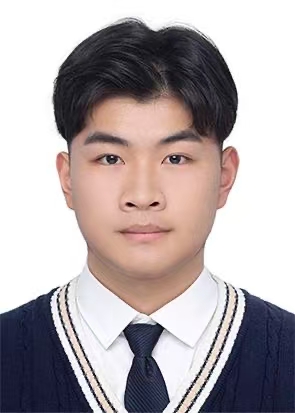}}]{Zedong Yang} is currently pursuing the Bachelor's degree in Automation at the School of System Design and Intelligent Manufacturing, Southern University of Science and Technology (SUSTech), Shenzhen, China. His research interests include deep learning, motion prediction, and the deployment of quadruped robots.
\end{IEEEbiography}

\newpage

\begin{IEEEbiography}[{\includegraphics[width=1in,height=1.25in,clip,keepaspectratio]{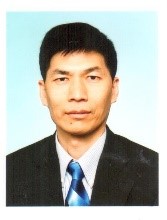}}]{Youfu Li}
  (Fellow, IEEE) received the PhD degree in robotics from the Department of Engineering Science, University of Oxford in 1993. From 1993 to 1995 he was a research staff in the Department of Computer Science at the University of Wales, Aberystwyth, UK. He joined City University of Hong Kong in 1995 and is currently professor in the Department of Mechanical Engineering. His research interests include robot sensing, robot vision, and visual tracking. In these areas, he has published over 400 papers including over 180 SCI listed journal papers. Dr Li has received many awards in robot sensing and vision including IEEE Sensors Journal Best Paper Award by IEEE Sensors Council, Second Prize of Natural Science Research Award by the Ministry of Education, 1st Prize of Natural Science Research Award of Hubei Province, 1st Prize of Natural Science Research Award of Zhejiang Province, China. He was on Top 2\% of the world’s most highly cited scientists by Stanford University, 2020, 2021 and Career Long. He has served as an Associate Editor for IEEE Transactions on Automation Science and Engineering (T-ASE), Associate Editor and Guest Editor for IEEE Robotics and Automation Magazine (RAM), and Editor for CEB, IEEE International Conference on Robotics and Automation (ICRA). He is a Fellow of the IEEE.
\end{IEEEbiography}	
\vspace{2mm}
\begin{IEEEbiography}[{\includegraphics[width=1in,height=1.25in,clip,keepaspectratio]{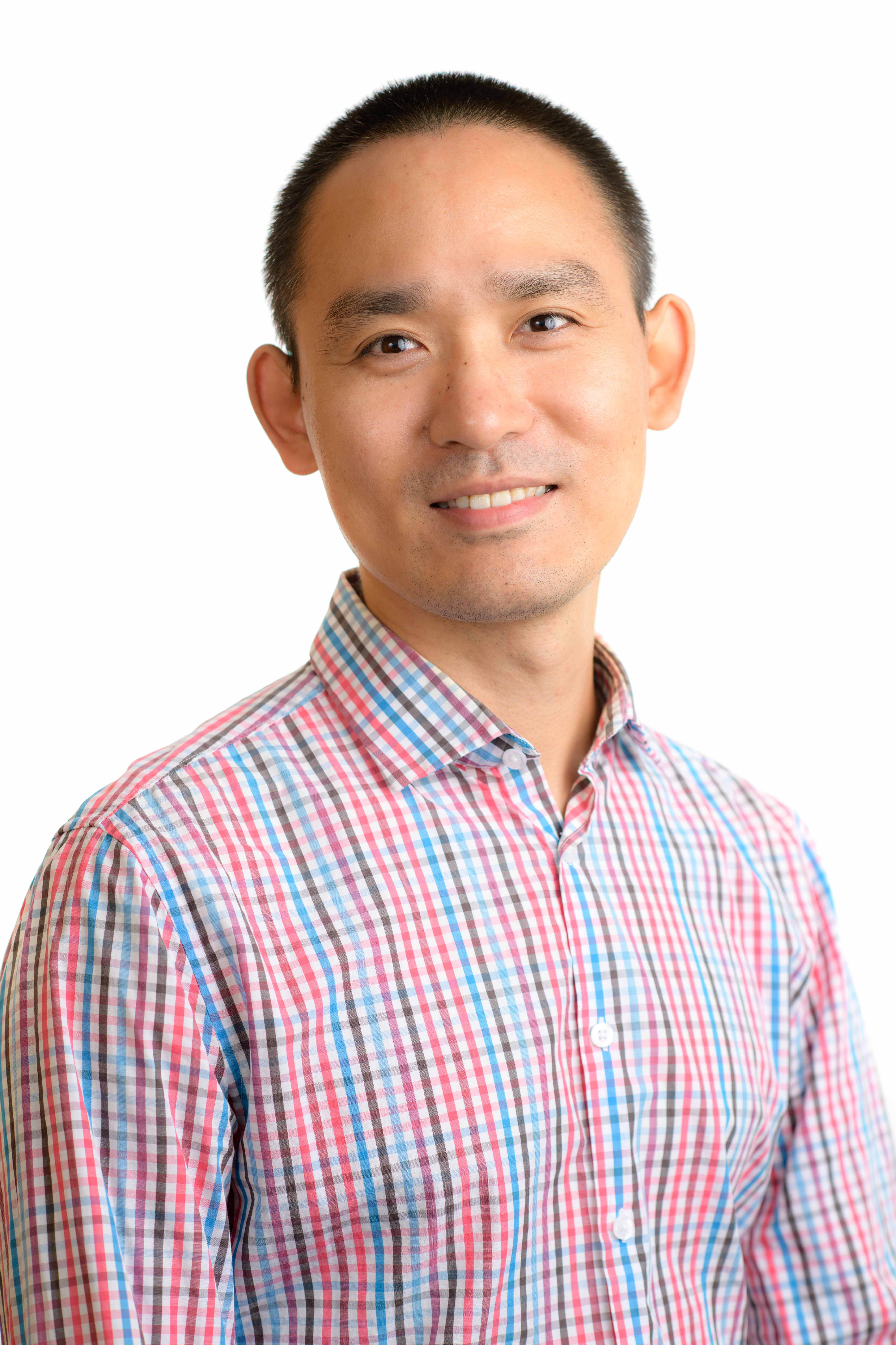}}]{He Kong}
  (Member, IEEE) received the Ph.D. degree in Electrical Engineering from the University of Newcastle, Australia, respectively. He was a Research Fellow at the Australian Centre for Field Robotics, the University of Sydney, Australia, during 2016–2021. In early 2022, he joined the Southern University of Science and Technology, Shenzhen, China, where he is currently an Associate Professor. His research interests include multi-modal perception, robot audition, state estimation, and control applications. He is currently serving on the editorial board of \textit{IEEE Robotics and Automation Letters}, \textit{IEEE Robotics and Automation Magazine}, \textit{IEEE Sensors Letters}, \textit{International Journal of Adaptive Control and Signal Processing}, etc. He has served as an Associate Editor on the IEEE Control System Society Conference Editorial Board as well as
  for several flagship conferences of the IEEE Robotics and Automation Society, including the
  IEEE ICRA, IEEE/RSJ IROS, the IEEE CASE. As a co-recipient, he has received the Best Paper Award at the 14th International Conference on Indoor Positioning and Indoor Navigation in 2024, the Best Poster Award at the 5th Annual Conference of China Robotics Society in 2024, etc.
\end{IEEEbiography}	

\vfill

\end{document}